%% file: main.tex
\documentclass{llncs}
\usepackage[figuresright]{rotating}
\usepackage{algorithmic}
\usepackage{algorithm}
\usepackage{amstext}
\usepackage{amsmath}
\usepackage[printonlyused]{acronym}
\usepackage{multirow,tabularx}
\usepackage{pgf}
\usepackage{tikz}
\usetikzlibrary{arrows,automata}
\usepackage{booktabs} % For formal tables
\usepackage{colortbl}

\usepackage{graphicx,fancyvrb}
\usepackage{latexsym}
\usepackage{amssymb}
\usepackage{amsmath}
\usepackage{amsfonts}
\usepackage{enumerate}
\usepackage{tikz}
\usetikzlibrary{shapes,shadows,calc,shapes.multipart}  
\usetikzlibrary{arrows,decorations.pathmorphing,backgrounds,fit,positioning,chains}
\tikzset{palla/.style={circle, shading=ball,ball color=black!40!blue!90, minimum width=0.5cm}}

\usepackage{multicol}
\usepackage[bookmarks=false]{hyperref}
\usepackage{pifont}
\input{comandi.tex}

\newcommand{\logic}{\mathsf{MeteoLOG}}
\newcommand{\meno}{-~}
\newcommand{\spaziatura}{~}
\newcommand{\RA}{AM{}}
\newcommand{\MDP}{LAM{}}
\newcommand{\HL}{\textsf{Tournament}}

\newcommand{\weathercondition}{numerical weather condition}
\newcommand{\weatherconditions}{numerical weather conditions}

\newcommand{\arpav}{ARPA Veneto}

\newcommand{\nega}{\neg}

\begin{document}

\title{``It could be worse, it could be raining'': reliable automatic meteorological forecasting for holiday planning}
%\dochead{22nd International Conference on Knowledge-Based and\\. Intelligent Information \& Engineering Systems}

\author{Matteo Cristani\inst{1} \and Francesco Domenichini\inst{2} \and Claudio Tomazzoli\inst{1} \and Luca Vigan\`o\inst{3} and Margherita Zorzi\inst{1} }

\institute{Dipartimento di Informatica, Universit\`a di Verona, Italy \and ARPAV, Italy \and Department of Informatics, King's College London, UK}

\maketitle

%\newcounter{example}
%\setcounter{example}{1}
%%%%%%%%%%%%%%%%%%%%%%%%%%%%%%%%%%%%%%%%%%%%%%%%%%%%%%%%%%%%%%%%%%%%%%%

\begin{abstract}
Meteorological forecasting provides reliable predictions about the 
%future 
weather within a given interval of time. %Meteorological forecasting can be viewed as a form of hybrid diagnostic reasoning and can be mapped onto an integrated conceptual framework. 
The automation of the forecasting process would be helpful in a number of contexts. For instance, %: when the amount of data is too wide to be dealt with manually; to support forecasters’ education; 
when forecasting about underpopulated or small geographic areas is out of the human forecasters' tasks but is central, e.g., for tourism. In this paper, we start to tame this challenging tasks: we  develop a defeasible reasoner for meteorological forecasting, which we evaluate on of a real-world example with applications to tourism and holiday planning.
% % We use the logic $\logic$, and the algorithm \HL, that produce a defeasible theory%$\logic$\ rests on several traditions, mainly on fuzzy, temporal and probabilistic logics. On this basis, we also introduce the algorithm \HL, that transforms a set of $\logic$\ rules into a defeasible theory for which an implemented automated deduction technology (called SpinDLE) exists. 
%  We evaluate the methodology by means of a real-world example of interest for the application to touristic business.

%Meteorological forecasting is the process of providing reliable prediction about the future weather within a given interval of time.
	%Forecasters adopt a model of reasoning that can be mapped onto an integrated conceptual framework. A forecaster essentially precesses data in advance by using some models of machine
	%learning to extract macroscopic tendencies such as air movements, pressure, temperature, and humidity differentials measured in ways that depend upon the model, but fundamentally, as
	%gradients. Limit values are employed to transform these tendencies in fuzzy values, and then compared to each other in order to extract indicators, and then evaluate these indicators
%	by means of priorities based upon distance in fuzzy values.
%	We formalise the method proposed above in a workflow of evaluation steps, and propose an architecture that implements the reasoning techniques.
\end{abstract}

\input{grafici_macro.tex}

\section{Introduction}
\label{sec:introduction}
\input{introduction.tex}

\vspace{-2ex}

\section{System architecture}
\input{architettura.tex}

%\section{Framework}
%\label{sec:frame}
%\input{framework.tex}

\vspace{-3.5ex}

\section{A defeasible reasoner for meteorological forecasting}
\label{sec:formalization}
\input{logic.tex}

\vspace{-2.5ex}

\section{Reference Implementation}
\input{referenceImpl.tex}

\input{SPINDLE_CONCLUSIONS.tex}

\vspace{-2.5ex}

\section{Related work}
\label{sec:related}

\vspace{-2.5ex}

\input{related}
\vspace{-2.5ex}

\section{Conclusions}

\vspace{-2.0ex}

\label{sec:concl}
\input{conclusions.tex}
%\input{SPINDLE_CONCLUSIONS.tex}

%Illustrazione metodo di sincronizzazione 
\bibliographystyle{abbrv}
\bibliography{bibliografia,bib2,biblioteometeo}
%\bibliography{bibliografia}      

\appendix

\newpage
\input{tournament.tex}

\input{referenceImpltwo.tex}
\input{SPINDLE_CONCLUSIONSTWO.tex}

\end{document}

%% file: comandi.tex
\newcommand{\comment}[1]{}

\def\makenewenum#1#2{%
\newcounter{cnt#1}
\newenvironment{#1}%
{\begin{list}{\makebox[0pt][r]{#2}}%
{\setlength{\itemsep}{0pt}%
 \setlength{\parsep}{.2em}%
 \setlength{\leftmargin}{2.5em}%
 \setlength{\labelwidth}{.2em}%
 \usecounter{cnt#1}}}%
{\end{list}}}
\makenewenum{enu}{\arabic{cntenu}.}
\makenewenum{enum}{\em(\arabic{cntenum})}
\makenewenum{itenum}{\em(\arabic{cntenum})}
\makenewenum{en}{(\roman{cnten})}
\makenewenum{renum}{\em(\roman{cntrenum})}
\makenewenum{enhyphen}{(\roman{cntenhyphen}')}

%% file: grafici_macro.tex
\newcommand{\fattoreScala}{.3}
\newcommand{\fattoreScalaMedio}{.55}
%
%%%%%%%%%%%%%%%%%%%%%%%%%%%%%%%%%%%%%%
%
% grafico con il concetto operativo
%

\newcommand{\graficoOperConc}[2]{
\begin{figure*}[!h]
\centering
\begin{tikzpicture}[scale= .4, transform shape]
\node[inner sep=0pt] (europe) at (-10,3)              {\includegraphics[width=.75\textwidth]{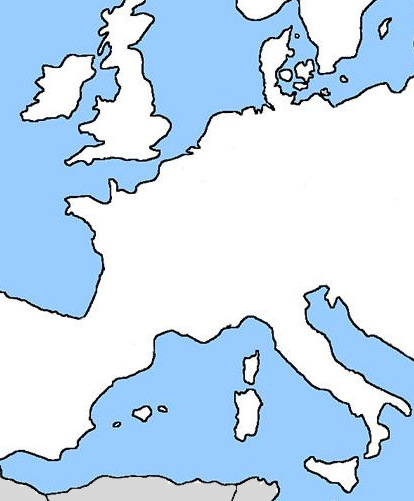}};
\node[palla,inner sep=0pt] (london) at (-13,6.2)              { };
\node[palla,inner sep=0pt] (rome) at (-8,-0.18)              { };
\node[inner sep=0pt] (muno) at (2,9)              {\includegraphics[width=.25\textwidth]{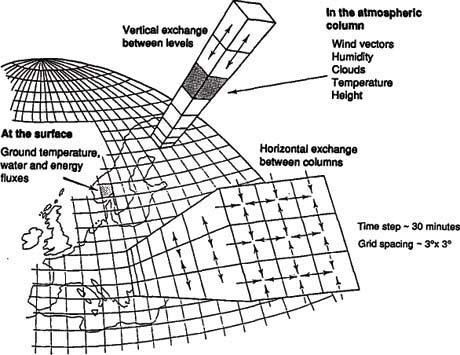}};
\node[inner sep=0pt] (mdue) at (6,9)              {\includegraphics[width=.25\textwidth]{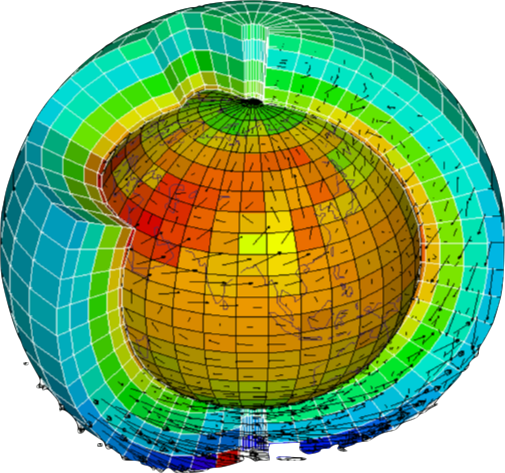}};
\node[inner sep=0pt] (mtre) at (10,9)                {\includegraphics[width=.25\textwidth]{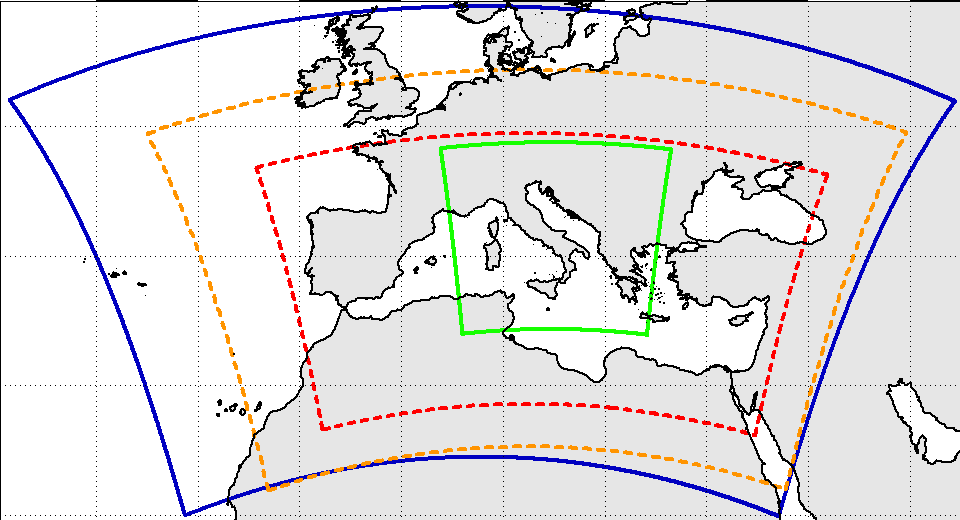}};

\node[inner sep=0pt] (elaboratore) at (6,3)     {\includegraphics[width=.15\textwidth]{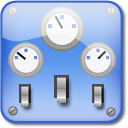}};
\node[inner sep=0pt] (output) at (12,3)           {\includegraphics[width=.42\textwidth]{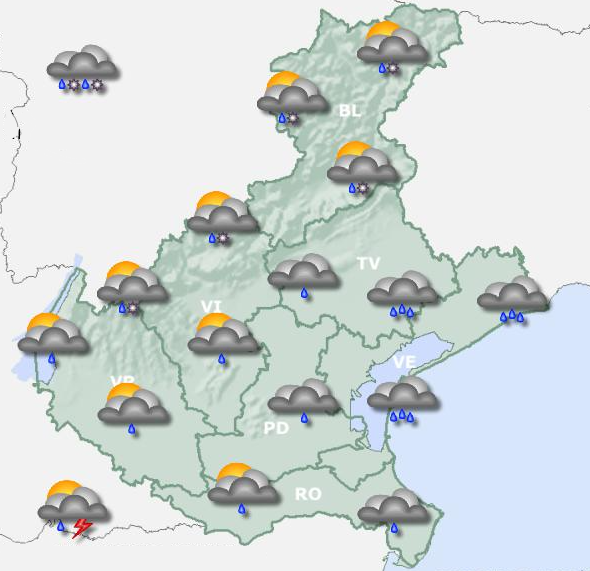}};
            
\draw[->,thin,dashed] (muno.south) -- (elaboratore.north);
\draw[->,thin,dashed] (mdue.south) -- (elaboratore.north);
\draw[->,thin,dashed] (mtre.south) -- (elaboratore.north);
\draw[->,thick] (rome.east) -- (elaboratore.west);
\draw[->,thick] (london.east) -- (elaboratore.west);
\draw[->,thick] (elaboratore.east) -- (output.west);

\node[inner sep=0pt] (baro) at (-7,4.5)              {\includegraphics[width=.30\textwidth]{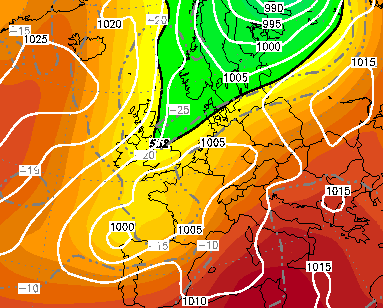}};
\node[inner sep=0pt] (mappaTemp) at (-2,0.5) {\includegraphics[width=.37\textwidth]{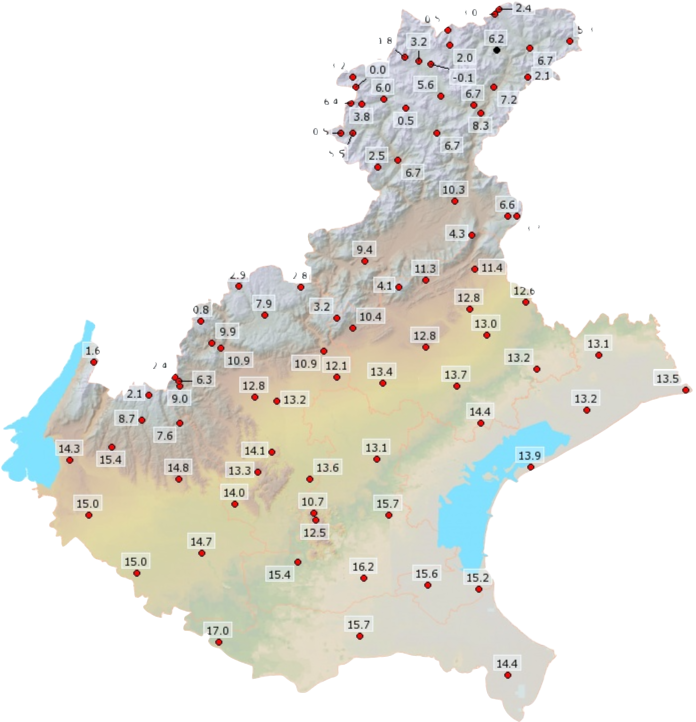}};

\end{tikzpicture}
\caption{#1} \label{#2}
\end{figure*}
}
%%%%%%%%%%%%%%%%%%%%%%%%%%%%%%%%%%%%%%
%
% grafico con il modello di architettura e dei moduli software
%
\newcommand{\graficoArchitettura}[2]{
\begin{figure*}[!h]
\centering
\begin{tikzpicture}[scale=\fattoreScalaMedio, transform shape]
 \node (c1) at (0,9) [minimum height=1cm,minimum width=3cm,draw, rounded corners, shade,  top color=orange!30, drop shadow] {Source Forecast Map};
 \node (c2) at (4.2,9) [minimum height=1cm,minimum width=3cm,draw, rounded corners,shade,  top color=orange!30, drop shadow] {Source Forecast Map};
 \node (t) at (2,7.5) [minimum height=1cm,minimum width=6cm,draw, rounded corners, shade,  top color=green!40, drop shadow] {Tournament};
 \node (shf) at (7,5) [minimum height=1cm,minimum width=4cm,draw, rounded corners, shade,  top color=blue!20, drop shadow] {Sharp Forecast};
  \node (smf) at (12,5) [minimum height=1cm,minimum width=4cm,draw, rounded corners, shade,  top color=blue!20, drop shadow] {Smooth Forecast};
  \node (r) at (2,1) [minimum height=1cm,minimum width=3cm,draw, rounded corners, shade,  top color=blue!40, drop shadow] {Reasoner};
  \node (o) at (2,5) [minimum height=2cm,minimum width=3cm,draw, rounded corners, shade,  top color=yellow!70, drop shadow] {Decision Maker};
  \node (b) at (16,5) [minimum height=1cm,minimum width=3cm,draw, rounded corners, shade,  top color=brown!90, drop shadow] {Bulletin generator};
  %\draw [fill=none,gray!10,very thin] (-3.5,-2) rectangle (8.5,3);
  \node (K)  at (-3,1) [cylinder, shape border rotate=90, draw,minimum height=4cm,minimum width=3cm, left color=blue!20, right color=blue!50] {Knowledge};
  \draw (c1.south) -- (t.north);
  \draw (c2.south) -- (t.north);
  \draw (t.south) -- (o.north);
  \draw (r.north) -- (o.south);
  \draw (r.north) -- (o.south);
  \draw (o.east) -- (shf.west);
  \draw (shf.east) -- (smf.west);
  \draw (smf.east) -- (b.west);
  \draw [gray!60,very thin,dashed](K.east) -| (r.west);
  \draw [gray!60,very thin,dashed](o.west) -| (K.north);
  \draw [gray!60,very thin,dashed](t.west) -| (K.north);
\end{tikzpicture}

\caption{#1} \label{#2}
\end{figure*}
}
%
%%%%%%%%%%%%%%%%%%%%%%%%%%%%%%%%%%%%%%
%
% grafico con il concetto operativo
%

\newcommand{\rerefenceImplGraph}[2]{
\begin{figure*}[!h]
\centering
\begin{tikzpicture}[scale=\fattoreScala, transform shape]
\node[inner sep=0pt] (emea) at (-15,3)              {\includegraphics[width=.75\textwidth]{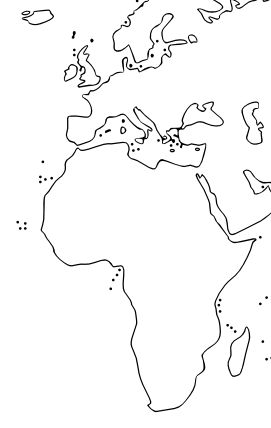}};
\node[inner sep=0pt] (europe) at (-4,3)            {\includegraphics[width=.75\textwidth]{./europe.png}};
\node[inner sep=0pt] (veneto) at (6,3)              {\includegraphics[width=.75\textwidth]{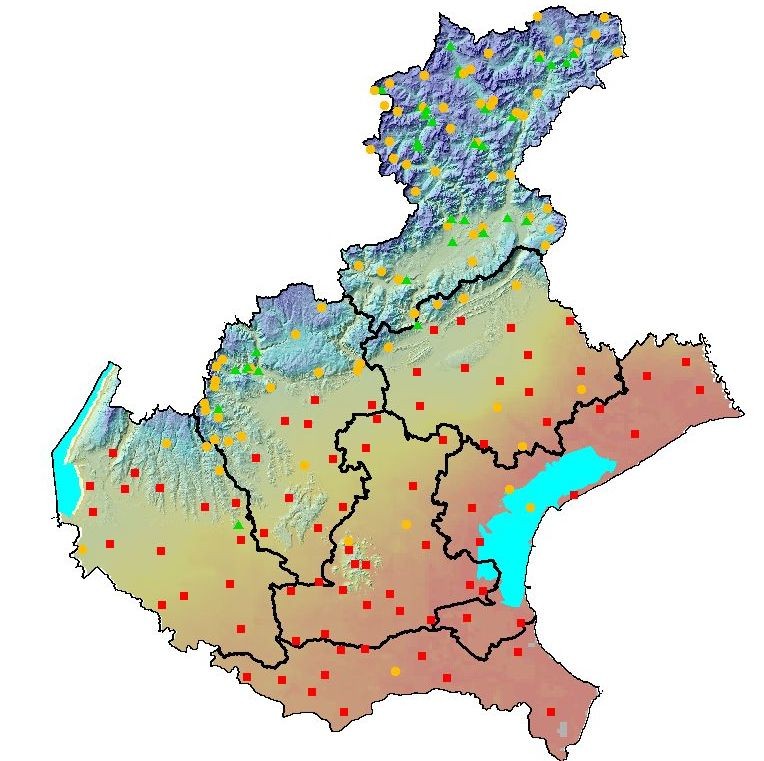}};
\node[inner sep=0pt] (lignano) at (16,8)              {\includegraphics[width=.75\textwidth]{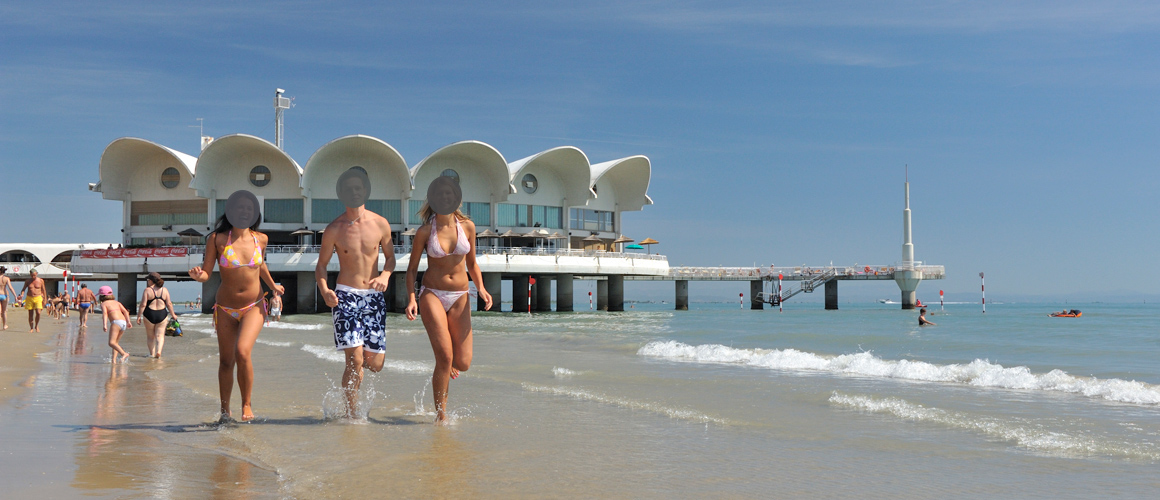}};
\node[inner sep=0pt] (venezia) at (16.5,2.5)              {\includegraphics[width=.65\textwidth]{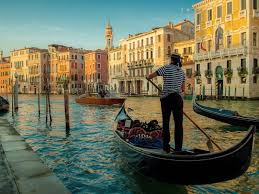}};
\node[inner sep=0pt] (chioggia) at (16,-3)              {\includegraphics[width=.75\textwidth]{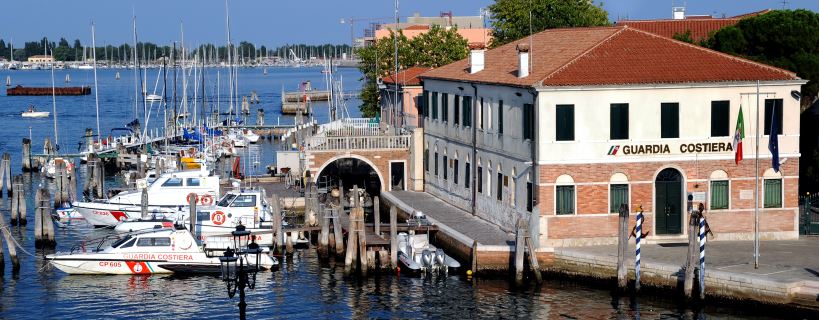}};

\node[circle, shading=ball,ball color=white!40!blue!90, minimum width=1.5cm,inner sep=0pt] (italia) at (-15,6)              { };
\node[circle, shading=ball,ball color=white!40!blue!90, minimum width=1.5cm,inner sep=0pt] (arpav) at (-3,1.5)              { };
\node[inner sep=0pt] (dummy) at (3,1.5)              { };
\node[circle, shading=ball,ball color=white!40!green!90, minimum width=1.5cm,inner sep=0pt] (nord) at (10,3)              {\Large{\textbf{N}}};
\node[circle, shading=ball,ball color=white!40!green!90, minimum width=1.5cm,inner sep=0pt] (centro) at (8,1)            {\Large{\textbf{C}}};
\node[circle, shading=ball,ball color=white!40!green!90, minimum width=1.5cm,inner sep=0pt] (sud) at (8,-1)              {\Large{\textbf{S}}};

\draw[->,thin,dashed] (italia) -- (arpav);
\draw[->,thin,dashed] (arpav) -- (dummy);
\draw[thin] (nord) -- (lignano.west);
\draw[thin] (centro) -- (venezia.west);
\draw[thin] (sud) -- (chioggia.west);
\end{tikzpicture}
\caption{#1} \label{#2}
\end{figure*}
}
%%%%%%%%%%%%%%%%%%%%%%%%%%%%%%%%%%%%%%
%
% tabella dei valori metereologici
%
\newcommand{\tabellaValoriMeteo}[2]{
\begin{figure}
{\scriptsize
\begin{center}
\begin{tabular}{clcclcclccl}
\ding{109} & \textbf{Snowfalls}:& \spaziatura &\ding{109}&\textbf{Wind}: & \spaziatura& \ding{109}&\textbf{Sea}: & \spaziatura & & \\
& \meno Blizzard & \spaziatura && \meno  Light Winds & \spaziatura & &\meno  Calm & \spaziatura & & \\
& \meno Snowstorm & \spaziatura &&\meno  Moderate Winds & \spaziatura &&\meno   Slight  & \spaziatura & & \\
& \meno Snow flurry & \spaziatura &&\meno  Moderate Winds & \spaziatura &&\meno  Moderate  & \spaziatura & & \\
& \meno Snow squall& \spaziatura &&\meno  Fresh Winds & \spaziatura &&\meno Rough  & \spaziatura & & \\
& \meno Snowburst & \spaziatura &&\meno  Near Gale & \spaziatura &&\meno  Very Rough & \spaziatura & & \\
& \meno Blowing snow & \spaziatura &&\meno Gale & \spaziatura &&\meno  High   & \spaziatura & & \\
& \meno Drifting snow  & \spaziatura &&\meno  Strong Gale & \spaziatura &&\meno  Very High  & \spaziatura & & \\
&& \spaziatura &&\meno  Storm  & \spaziatura & & \meno  Phenomenal   & \spaziatura & & \\
& & \spaziatura &&\meno  Violent Storm & \spaziatura && & \spaziatura & & \\
& & \spaziatura && & \spaziatura && & \spaziatura & & \\
\ding{109} & \textbf{Sky conditions}:& \spaziatura &\ding{109}&\textbf{Precipitation}: & \spaziatura& \ding{109}&\textbf{Rainshowers}: & \spaziatura& \ding{109}&\textbf{Visibility}:\\
& \meno Clear or Sunny Skies& \spaziatura && \meno No precipitation & \spaziatura & &\meno  Scattered & \spaziatura & &\meno  Clean \\
& \meno Partly Cloudy & \spaziatura &&\meno  Very Light Rains & \spaziatura & &\meno  Isolated & \spaziatura & &\meno  Misty \\
& \meno Mostly Cloudy & \spaziatura &&\meno  Light Rains & \spaziatura & &\meno  Occasional  & \spaziatura & &\meno   Foggy \\
& \meno Cloudy & \spaziatura &&\meno  Moderate Rains & \spaziatura & &\meno  Squally  & \spaziatura & &\meno  Hazy \\
& \meno Overcast & \spaziatura &&\meno  Heavy Rains & \spaziatura & &\meno  & \spaziatura & & \\
\end{tabular}
\caption{#1}\label{#2}
\end{center}	
}
\end{figure}
}

%% file: introduction.tex
In the last ten years or so, meteorological forecasting has become a commonly required web service and meteorological forecasting websites are nowadays one of the most expensive websites for advertisement. Producing a meteorological forecast is, however, an expert task to perform. It is a human activity that is typically provided as a partially automated pipeline in which a first step consists in generating \emph{models} of the evolution of the weather in a given geographic area. 
These models are often not directly accessible online because of their sizes, so the forecaster relies on the execution of a sophisticated reasoning on the data, comparing the models and evaluating the confidence degree in a range of possibilities compatible with the models themselves.

% and the ways in which this typically human high-level activity is provided as a partially automated pipeline in which a first step consists in generating \emph{models} of the evolution of the weather in a given geographic area. 
% The forecaster, once received these models, usually available, but not directly accessible online because of their sizes, execute a sophisticated reasoning with the data themselves, compare the models and evaluate its confidence degree in a range of possibilities compatible with the models themselves.

Although this process is an expert one, and it is driven by heuristic knowledge of the domain, there exist some traits of the reasoning method that are systematic, and used in force of the nature of the domain itself. In other words, there exists a specific form of \emph{meteorological reasoning} consisting in the pipeline described above, and possibly in other collateral steps including cmpaison with real world data against models, and local methods used to provide the inference as terms of comparison in the range itself. 

The computer technologies currently available don't satisfy the requirements of \emph{quality} in general application configurations. In this paper, we investigate how to simulate the behavior of a meteorologist in this processes, in an extended way with respect to what has been done in various previous studies, and specifically inspired by the approach followed by Ramos-Soto et al.~\cite{Ramos-Soto201544}.

In particular the current technologies suffer from a number of severe drawbacks: (i)~They are based on the crisp interpretation of one single model so that the \emph{reliability} depends on that model;  (ii)~They refer to vast areas and tend to infer forecasting of smaller areas without considerations of the local effects;
(iii)~They sometimes include extreme simplifications of the forecast itself, providing information in graphical terms and thus unifying complex judgments into single ones.

It therefore make senses to envision an automated expert technology able to support or even substitute the forecaster in many use cases, including:
(i)~ \textit{As a learning tool:} when an expert forecaster helps students, or newbies at the workplace, to become experts of the pipeline mentioned  above, it would be beneficial to provide a learning environment simulating \emph{reasonably} correct forecasts.
(ii)~ \textit{As a decision support system:} comparing human decisions where the level of confidence is not particularly high with decisions made by an AI tool would help the forecaster in providing a better forecast.
(iii)~ \textit{When forecasting is not sustainable:} an AI tool may substitute entirely humans when the requested forecasting is too finely associated with territory portions.
The latter is the main motivation behind our research. 

We aim at defining an AI tool for meteorological forecasting that could be applied to real-world situations in which the tourism flow is not intense enough to make the employment of human forecasters acceptable from a business viewpoint, whilst an automated tool, which provides the forecasting, albeit possibly in a not particularly accurate way, would be beneficial.

We use non monotonic reasoning and make thus possible to accommodate conflicting rules in the system, so that the process of decision making not only results fuzzy, but is also subject to revisions and constrained by confidence;
To better explain what we focus upon, consider the following example of a specific situation managed by the framework that we devise here.
\begin{example}
Let us consider a remote area, such as a small island in the middle of the Pacific Ocean. This island has a limited tourism flow and it is thus unlikely that forecasting by authorities on the island itself would be provided. Hence, the level of forecasting is derived by mid-scale models on the Ocean that are very unreliable for this and other small islands.
In this island, however, there is a satellite high-speed Internet connection and thus a web server can be installed. Moreover, on the island's coasts and its sole mountain, there exist a group of wind 
and rain sensors connected to the server through a network. It is therefore possible to install a sophisticated  technology that makes a forecast available through a web service to the tourists interested in visiting the island.
\end{example}

%\fix{Luca}{Questo \`e un ottimo esempio, ma poi invece nel paper parliamo del Veneto (anche se non chiariamo bene il vantaggio della presenza di Francesco) e questo sembra contraddire quanto appena detto, visto che il Veneto non \`e un posto sperduto con poche informazioni. Questa va chiarito molto bene, e al limite bisogna togliere questo esempio, oppure spiegare molto molto bene nelle prossime sezioni perche\'e quello che discutiamo \`e rilevante sia in general sia per l'isoletta}

% \textbf{Synopsys.} 
We proceed as follows. In Section~\ref{architecture} we introduce the architecture of our framework. In Section~\ref{sec:formalization} we describe the logic and the algorithm we use to produce input rules for a reasoner that will derive a \emph{weather scenario}; we also describe the steps behind the release of the final weather bulletin. Section~\ref{refimpl} is devoted to a real-world example. 
In Section~\ref{sec:related} we discuss related work, and in Section~\ref{sec:concl} we draw conclusions.

%% file: architettura.tex
\label{architecture}

\graficoOperConc{The operative concept of the system}{fig:operconc}

The system aim is to produce a better weather forecast, given meteorological models and data gathered from the field, as summarized in Figure \ref{fig:operconc}.
Figure~\ref{fig:Modules} shows the logic model of the system architecture, which comprises several modules, each one with a single responsibility as described below.

%Here the description of each module:
%Every module is related to at least one other, while all refer to one named \textit{Orchestrator}:
\begin{description}
\item [Source Forecast Map:] this module aims at the retrieval of raw informations form a specific source (i.e., Temperature, Humidity, ...) connecting to a sensor network or a data source. To add a new source to the system (e.g., Sea Status) one will need to extend only the implementation of this module. The output of this module is a ``source forecast map''

\item [Tournament:] this module takes as input source forecast maps and their \emph{accuracy} and \emph{fragility} data, and gives \emph{forecast rules} to be examined by the \textit{Reasoner}; see Section~\ref{algorithm} for an informal description of the algorithm.

\item [Decision Maker:] this module decides which model will give best performances and is thus the actual core of the system.

\item [Reasoner:] this module is the ``brain'' of the system as it applies our deduction system to make decisions about which forecast draft is the ``best one''.

\item [Knowledge base:] this is where the knowledge base is stored; the \textit{Reasoner} will access it for reasoning and the \textit{Decision Maker} will increment it after the evaluation of the results of the reasoner. 

\item [Sharp Forecast:] this module provides a mapping from quantitative forecasting (numerical) to qualitative forecasting (words); see Section~\ref{sec:sharpforecast} for details.

\item [Smooth Forecast:] this module transforms the forecasting  expressed in words into natural language sentences suitable to be delivered to the public; see Section~\ref{sec:smoothforecast} for details.

\item [Bulletin generator:] this module provides visualization of all data, building the output of the system as a ``pdf'' file or a hypertext.
\end{description}
\graficoArchitettura{Logic model of system architecture}{fig:Modules}

%% file: logic.tex
What is commonly intended as ``weather forecasting'' can be logically model as a \emph{conclusion} the forecaster derives from a set of \emph{premises}, by the application of some (both deductive and empirical) rules.%\fix{Luca}{Questa \`e una spiegazione chiara e semplice che va messa prima! La prima frase forse gi\`a nell'introduzione e la seconda all'inizio di questa sezione, per spiegare quello che facciamo}
%
% \footnote{An extend version of this work is available online at \ARXIV}
\subsection{The logic $\logic$}
The reasoning process of the weather forecaster is formally built upon technical steps implementing the workflow described in Section \ref{architecture}. In this section, we develop a logical framework called $\logic$, that formalizes the hybrid reasoning at the basis of meteorological forecasting. 
$\logic$, informally introduced in~\cite{DBLP:conf/kes/CristaniDOTZ18}, benefits from three standard logical approaches: defeasible logic~\cite{Governatori2014168}, labeled deduction systems~\cite{MVZ-jmvl,ViganoVZ14,ViganoVZ17,cotz18} and fuzzy/non-deterministic/probabilistic frameworks~\cite{Domanska20127673,AZ13}.
%\textcolor{blue}{va bene lasciare il riferimento, ma non vedo cosi evidente la relazione con fuzzy, allo stato attuale}
%The main idea is to represent the heuristic knowledge described by the the previously defined architecture, that couples deductive, empirical and fuzzy reasoning.%forecasting 

We introduce the \emph{syntax of formulae} and of \emph{labels}, along with a notion of \emph{prevalence}, which imports a defeasible flavour into the system. We also provide an intuitive description of the label-elimination algorithm \HL, which represents the basis of the reasoning process that we develop below. %The full procedure \HL{} is described in Section\ref{}.

In this paper, we only deal with ground formulas modeling meteorological forecasting values, which we will call \emph{Assertional Maps (\RA{s})}. 
\RA{s} provide quantitative information and they represent the basic piece of knowledge used for forecasting. In the real world, they are collected worldwide, from different forecasting sites and through a number of different technologies. The internationally accepted set of \weatherconditions\ revealed in \RA{s}\ concerns:
% the following concepts/quantities: 
\emph{Temperature}, \emph{Pressure}, \emph{Humidity}, \emph{Snowfalls}, \emph{Wind}, \emph{Precipitations}, \emph{Visibility}.
 
From an abstract viewpoint, \RA{s} express rough assertions about weather to be processed and evaluated. They are simply represented by suitable predicates on space-time coordinates pointing out a \weathercondition, expressed in a suitable measuring system.
	 In particular,  the temperature is expressed in graders, pressure  is expressed in \emph{HPa}, humidity   in \emph{percentage}, rains  are expressed  in \emph{millimetres}, snowfalls   in \emph{centimetres}, visibility  is expressed in \emph{metres}, Wind  in \emph{Knots},  Cloudiness (C) in \emph{percentage}  . 

% 
% \RA{s} provide quantitative information and they represent the basic piece of knowledge that we deal with. In the real world, they are collected worldwide, from different forecasting sites and through a number of different technologies.  
% The internationally accepted set of \weatherconditions\ revealed in \RA{s}\ concerns:
% % the following concepts/quantities: 
% \emph{Temperature}, \emph{Pressure}, \emph{Humidity}, \emph{Snowfalls}, \emph{Wind}, \emph{Precipitations}, \emph{Visibility}.
%
Formally, an \RA{} is a five-ary predicate 
% five-dimensional predicates of the shape 
$Q(x,y,z,\tau^{r},q)$, where Q is a \weathercondition, $x$, $y$ and $z$ represent geographic coordinates, $\tau^{r}$ represents the forecasting time (the interval of the validity of the assertion) and $q$ is the effective revealed value, represented by a 2-dimensional vector $(v,d)$, where $v$ is the numerical value and $d$ is the \emph{direction}.
For instance, $\mathit{Rain}(45.43, 11.80, 06/04/2018, \mbox{14:05:00}\mathit{CET}, 5\mathit{mm})$ represents that the assertion $\mathit{Rain}$ on ground level, point of measure $(45.43,11.80)$ on GPS coordinates, on $06/04/2018$ at $\mbox{14:05:00}$ $\mathit{CET}$ was 5 millimetres.

% \RA{s} provide quantitative information and they represent the basic piece of knowledge that we deal with. In the real world, they are collected worldwide, from different forecasting sites and through a number of different technologies.  

Since we are interested in the reasoning process behind the forecasting, we now focus on \emph{models} experts apply to derive information from \RA. We formalize such methods and related notions by means of labels, and import into the formulas additional information such as the precision of the method and the detection time (the instant in which the method has been applied to generate the map). This information is crucial for the forecaster's work, since the choice of the (as much as possible) correct maps is mainly based on methodological information.

\emph{Labeled Assertional Maps (\MDP{s})} are obtained by labelling \RA{s}. This formally models the additional information the forecaster have to evaluate and decide if a rough \RA{} expressing a prediction is admissible for forecasting or not.

Labels represent \emph{contextualised methods}, i.e., a forecasting method applied to a data gathering sample, performed in a given instant of time, weighted with an some accuracy information; they are pairs of the kind $\langle \lambda, \tau^{t}\rangle$, where $\lambda$ represents a model and $\tau$ represents the instant in which the map has been generated.

Each method can be associated with an accuracy value $\lambda.a$, a function that extracts the accuracy information from the method $\lambda$.

An \MDP{} is then a labeled formula $\langle \lambda, \tau^{t}\rangle:Q(x,y,z,\tau^{r},q)$. 

Labeled ground formulas that express \weatherconditions\  permits to compare different formulas expressing the same forecasting concept on the basis of different methods and time. Some  priority rules allow us to decide what set of  sources is the more reliable one. As explained  in the following, we will use some priority rules to order \RA s, in order to eliminate the ones that do not overcome a given threshold of reliability.
	 
%\paragraph{Prevalences between \MDP}. Following 
We introduce now some relation between labels. This step also imports  a defeasible behaviour in the system.% With MCA-priorities we import  a defeasible behaviour, by stating some priority relations between labeled assertions. %Priority  relates the forecasting conclusions of a model with the ones of another when one of the models is considered for that particular 
	 %variable more likely to forecast correctly than the other one:
	 There are two main kind of priority relation. The first one, called here \emph{quantitative} or \emph{algorithmic} { priority}, simply automatically check and compare labeled assertions on the basis of their quantitative information (accuracy and time).
\noindent	
$\langle \lambda_1,  \tau_1\rangle :Q_1(x,y,z,\tau^{r},q_1) \succ \langle  \lambda_2, \tau_2\rangle :Q_2(x,y,z,\tau^{r},q_2)$, with $Q_1=Q_2$ holds in one of the following cases:
	
	\begin{enumerate}
	\item $\lambda_1.a>\lambda_2.a$;
	%\item $a(\lambda_1)=a(\lambda_2)$ and $\epsilon_1\leq\epsilon_2$;
	\item $\lambda_1.a=\lambda_2.a$ and $\tau_1<\tau_2$ 
	%\item $a_1\geq a_2 \wedge (\tau-\tau_1)>(\tau-\tau_2)$  
	%\item $a_1< a_2 \wedge (\tau-\tau_1)>(\tau-\tau_2)\wedge \epsilon_1>\epsilon_2$
\end{enumerate}	 
	
%	\textcolor{blue}{Dobbiamo stabilire tutte le priorit\`a} 

Notice that more accurate model-based assertions always (quantitatively) prevails, and, up to equal accuracy, prevails the  most recent \MDP.% and data error.

Given a set of \MDP, each assertional maps is weighted according to the relations described above and the whole set is ordered as a consequence.

%\begin{example}[Algorithmic  Prevalence between \MDP]
%\end{example}

The second kind of priority relation is called \emph{experience-based} or \emph{specific}, and models the empirical knowledge of the expert of the domain.

Following the real world setting, we state that a specific priority always prevails on an algorithmic one (w.r.t. the same assertion).

Having defined the syntax of $\logic$, we are currently working at the definition of a suitable semantics and a natural deduction system~\cite{Prawitz65,AZ16} for $\logic$.

\subsection{The \HL{} algorithm}
\label{algorithm}
\HL{} is an algorithm whose aim is  to return a  a defeasible theory, given an ordered set of \emph{Metarules}, a set of \emph{accuracies} and actual time.
Informally, the algorithm maps $assertions$ into defeasible rules and facts; when it finds possible conflicts it generates a set of defeasible conflicting rules and then, using the accuracy information, it generates the priority rules to solve the conflict so that the method with best accuracy prevails; in case of even accuracy, the latest \MDP{} prevails. 
%It takes in input a set of \emph{accuracies} of the methods $\Psi$, the current time $\mathsf{t}$, an ordered set of \emph{Metarules} $\Theta$, where metarules are the previously defined \MDP{} in the form $\mathit{label : assertion}$ where labels are of the form $\langle \mathit{method,time} \rangle$ and $\mathit{assertions}$ are of the form $name(position,time,value)$;\fix{Luca}{Scusate ma questa \`e una pessima frase, che sembra contraddire/ripetere quanto detto sopra. Va pulito e semplificato} its output is a defeasible theory $\cal{T}$ $= \langle F, R, P \rangle$ (facts, rules, priorities).

%\fix{Luca}{Qui serve una frase di introduzione, che dica cosa sia \HL{} e in che senso \`e un nostro contributo. Mi sembra che questa subsection vada riscritta completamente, sia come flusso (invertendo quello corrente che prima va nel dettaglio e poi da spiegazioni) sia come formalizzazione}
Once a set of \MDP{s} has been collected and an \emph{accuracy} set has been acknowledged, our \HL{} algorithm starts with a sifting action on the set of labeled assertions. 
First, it discharges \MDP{s} that are out of date.
Second, it orders \MDP{s} on the basis of priorities, obtaining some \RA{} for each \weathercondition\ we are interested in. This operation corresponds to a label-elimination: once priorities have been derived, the majority of information about the forecasting method became useless.  
As an output of this step, we obtain a set of \emph{defeasible rules} to be given as input to the \emph{Reasoner}, which will derive a set of \weathercondition\ also called a weather \emph{scenario}.
Priorities of these generated defeasible conflicting rules are given by a \emph{function} (named ``supremacy'') that takes into account two conflicting rules so that, depending on this function definition, the resulting output can differ from both source rules. 
The pseudocode of the \HL\ algorithm is in Appendix \ref{algorithm}
\vspace{-3ex}

\subsection{Weather Forecasting Reasoning}
\vspace{-1ex}
Forecasting reasoning can be divided into three steps: (i) the quantitative forecasting (invisible to the final user) the reasoner generates; (ii) the qualitative forecasting (also invisible to the final user) called in the following \emph{sharp forecasting}; i(ii) the qualitative, natural language based forecasting destined to the final user, called in the following \emph{smoothed forecasting}. Between the first two phases, a mapping between data and a suitable \emph{forecasting lexicon} occurs.

\subsubsection{From data to pre-bulletin: Sharp Forecasting}
\label{sec:sharpforecast}
Once the final set of reliable assertional maps has been collected, the forecaster can proceed with data analysis and the releasing of the weather bulletin.
%
%The conclusion regards a weather condition and premises are anything else that assertions about physical quantities. 
%
Meteorological conditions represent a crucial step of the forecasting reasoning. The internationally accepted ranges for meteorological conditions are shown in Figure~\ref{fig:scales}. To each value of the range a precise interval in a suitable measure scale can be associated.

\tabellaValoriMeteo{Metereological Conditions}{fig:scales}

\subsubsection{The weather forecasting lexicon: smoothed forecasting}
\label{sec:smoothforecast}
Every one has a wide experience in weather forecasting as a final user. It is well known that weather bulletins are offered in a friendly form. For example, if the forecaster deduces that the probability that tomorrow it will rain is very small, she doesn't release the assertion ``$\mathit{Rain}(45.43, 11.80, 06/04/2018, \mbox{14:05:00}\mathit{CET}, 5\mathit{mm})$ with probability $15\%$''  but the understandable natural language sentence ``partially cloudy, possible scattered rains''. This final step provides a ``smoothing'' phase to the output of the previous one in which some adjectives can be added to give evidence to the uncertainty of the event. We don't fully model this final step of weather forecasting; nonetheless, in our reference implementation we propose an example of a possible automatisation of the human task, leaving the full development for future work (see Section~\ref{sec:concl}).

%% file: referenceImpl.tex
\label{refimpl}

To illustrate concretely how our approach can fit a real-life scenario, let us consider a weather forecast considering the seaside part of Veneto, our region, which is located in the north-east of Italy and albeit being not a remote area it exploits several neighbor small touristic places.\rerefenceImplGraph{Some of the touristic places in our region}{fig:cartina}
For the sake of space, but without loss of generality, we limit the weather forecast to cloud, wind and sea conditions and to only three points; we label these points North, South and Center, the latter representing roughly the position of the famous city of Venice (see Figure~\ref{fig:cartina}). Sea Conditions have only one point, representing the sea in the area. We use only two forecasting maps and we limit the time-frame to only two values, representing two and one days after the present: respectively $t_2,t_1,t_0$
We have as input two forecasting sources, coming from different forecasting models such as IFS (also known as ECMWF for European Center Medium Weather Forecast) and GFS (Global Forecast System), plus the map of observations.

The first source obtained with the GFS prevision model asserts, using $N$ for $form~ North$ and $E$ for $form~ East$,at time $t_0$\\
{\small 
\begin{tabular}{l}$North:\{cloudiness: 90\%, Wind:18 knots ~ N\}$ \\$Center:\{cloudiness: 90\%, Wind:18 knots ~ N \}$\\$South:\{cloudiness: 90\%, Wind:10 knots ~ N,\} Sea: 190 cm~ wave $\\\end{tabular} \\  at time $t_1$   \\\begin{tabular}{l}$North:\{cloudiness: 90\%, Wind:8 knots ~ N\}$ \\$Center:\{cloudiness: 90\%, Wind:8 knots ~ E \}$ \\$South:\{cloudiness: 90\%, Wind:5 knots ~ E,\} Sea: 100 cm~ wave $\\\end{tabular} \\ at time $t_2$   \\\begin{tabular}{l}$North:\{cloudiness: 90\%, Wind:8 knots ~ N\}$\\ $Center:\{cloudiness: 90\%, Wind:8 knots ~ E \}$ \\$South:\{cloudiness: 90\%, Wind:5 knots ~ E,\} Sea: 100cm~ wave $
\end{tabular}
}

The second source obtained with ECMWF  asserts, at time $t_0$ \\
{\small 
\begin{tabular}{l}$North:\{cloudiness: 90\%, Wind:15 knots ~ NE\}$ \\$Center:\{cloudiness: 90\%, Wind:15 knots ~ NE \}$\\$South:\{cloudiness: 90\%, Wind:15 knots ~ NE,\} Sea: 160 cm~ wave $\\\end{tabular} \\  at time $t_1$ \\\begin{tabular}{l}$North:\{cloudiness: 75\%, Wind:5 knots ~ NE\}$ \\$Center:\{cloudiness: 75\%, Wind:5 knots ~ NE \}$ \\$South:\{cloudiness: 75\%, Wind:5 knots ~ N,\} Sea: 90 cm~ wave $\\\end{tabular} \\  at time $t_2$ \\\begin{tabular}{l}$North:\{cloudiness: 30\%, Wind:5 knots ~ N\}$ \\$Center:\{cloudiness: 30\%, Wind:5 knots ~ N \}$ \\$South:\{cloudiness: 30\%, Wind:5 knots ~ N,\} Sea: 50~ wave $
\end{tabular}
} 

The observation map, which only relates data at  $t_0$ states that\\
{\small 
\begin{tabular}{l}$North:\{cloudiness: 90\%, Wind:15 knots ~ NE\}$ \\$Center:\{cloudiness: 90\%, Wind:15 knots ~ NE \}$ \\$South:\{cloudiness: 90\%, Wind:15 knots ~ NE,\} Sea: 190 cm~ wave $
\end{tabular}
}

We know from knowledge experts that ECMWF has a better accuracy than GFS: numerically $a(ECMWF,t_1)=0.85$, $a(ECMWF,t_2)=0.80$ , $a(GFS,t_1)=0.45$, $a(GFS,t_2)=0.40$. 
These assertions, using ``E'' for ECMWF,  ``G'' for GFS, ``O'' for observation and  ``C'' for ``cloudiness'', ``W'' for ``wind'' and ``S'' for ``sea conditions'' can be represented in our formalism as
\begin{center}
{\scriptsize 
\begin{tabular}{lll}
$\langle G,t_0 \rangle :C(North,t_0,90)$&$ \langle G,t_0 \rangle :C(Center,t_0,90)$&$ \langle G,t_0 \rangle :C(South,t_0,90)$\\
$\langle G,t_0 \rangle :C(North,t_1,90)$&$ \langle G,t_0 \rangle :C(Center,t_1,90)$&$ \langle G,t_0 \rangle :C(South,t_1,90)$\\
$\langle G,t_0 \rangle :C(North,t_2,90)$&$ \langle G,t_0 \rangle :C(Center,t_2,90)$&$ \langle G,t_0 \rangle :C(South,t_2,90)$\\
$\langle E,t_0 \rangle :C(North,t_0,90)$&$ \langle E,t_0 \rangle :C(Center,t_0,90)$&$ \langle E,t_0 \rangle :C(South,t_0,90)$\\
$\langle E,t_0 \rangle :C(North,t_1,75)$&$ \langle E,t_0 \rangle :C(Center,t_1,75)$&$ \langle E,t_0 \rangle :C(South,t_1,75)$\\
$\langle E,t_0 \rangle :C(North,t_2,50)$&$ \langle E,t_0 \rangle :C(Center,t_2,50)$&$ \langle E,t_0 \rangle :C(South,t_2,50)$\\
$\langle O,t_0 \rangle :C(North,t_0,90)$&$ \langle O,t_0 \rangle :C(Center,t_0,90)$&$ \langle O,t_0 \rangle :C(South,t_0,90)$\\

$\langle G,t_0 \rangle :W(North,t_0,[N,18])$&$ \langle G,t_0 \rangle :W(Center,t_0,[N,18])$&$ \langle G,t_0 \rangle :W(South,t_0,[N,10])$\\
$\langle G,t_0 \rangle :W(North,t_1,[N, 8])$&$ \langle G,t_0 \rangle :W(Center,t_1,[E,8])$&$ \langle G,t_0 \rangle :W(South,t_1,[E,5])$\\
$\langle G,t_0 \rangle :W(North,t_2,[N,8])$&$ \langle G,t_0 \rangle :W(Center,t_2,[E,8])$&$ \langle G,t_0 \rangle :W(South,t_2,[E,5])$\\
$\langle E,t_0 \rangle :W(North,t_0,[NE,15])$&$ \langle E,t_0 \rangle :W(Center,t_0,[NE,15])$&$ \langle E,t_0 \rangle :W(South,t_0,[NE,15])$\\
$\langle E,t_0 \rangle :W(North,t_1,[NE,5])$&$ \langle E,t_0 \rangle :W(Center,t_1,[NE,5])$&$ \langle E,t_0 \rangle :W(South,t_1,[NE,5])$\\
$\langle E,t_0 \rangle :W(North,t_2,[N,5])$&$ \langle E,t_0 \rangle :W(Center,t_2,[N,5])$&$ \langle E,t_0 \rangle :W(South,t_2,[N,5])$\\
$\langle O,t_0 \rangle :W(North,t_0,[NE,15])$&$ \langle O,t_0 \rangle :W(Center,t_0,[NE,15])$&$ \langle O,t_0 \rangle :W(South,t_0,[NE,15])$\\

$\langle G,t_0 \rangle :S(Sea,t_0,190)$&$ \langle G,t_0 \rangle :S(Sea,t_1,100)$&$ \langle G,t_0 \rangle :S(Sea,t_2,100)$\\
$\langle E,t_0 \rangle :S(Sea,t_0,160)$&$ \langle E,t_0 \rangle :S(Sea,t_1,50)$&$ \langle E,t_0 \rangle :S(Sea,t_2,10)$\\
$\langle O,t_0 \rangle :S(Sea,t_0,190)$& &
\end{tabular}
}
\end{center}
This is therefore our set of metarules, so after the \emph{Translator} has done its elaboration using algorithm described in \ref{algorithm} we can have:
\begin{center}
{\scriptsize
\begin{tabular}{llllllllllllllllllllllll}
\multicolumn{3}{l}{$r_{fcg_{11}}: $}& \multicolumn{3}{l}{ $ \Rightarrow CN_g{t_0}90$ }~~~~~~~& %~~~~~~~~~~ &
\multicolumn{3}{l}{$r_{fcg_{21}}: $}& \multicolumn{3}{l}{ $ \Rightarrow  CN_g{t_1}90$}~~~~~& %~~~~~&
\multicolumn{3}{l}{$r_{fcg_{31}}: $}& \multicolumn{3}{l}{ $ \Rightarrow  CN_g{t_2}90$}~~~~~&%~~~~~&
\multicolumn{3}{l}{$r_{co_{11}}:  $}& \multicolumn{3}{l}{ $ \rightarrow  CN{t_0}90$}\\
\multicolumn{3}{l}{$r_{fcg_{12}}: $}& \multicolumn{3}{l}{ $ \Rightarrow CC_g{t_0}90$}& %~~~~~~~~~~ &
\multicolumn{3}{l}{$r_{fcg_{22}}: $}& \multicolumn{3}{l}{ $ \Rightarrow  CC_g{t_1}90$}& %~~~~~&
\multicolumn{3}{l}{$r_{fcg_{32}}: $}& \multicolumn{3}{l}{ $ \Rightarrow  CC_g{t_2}90$}& %~~~~~&
\multicolumn{3}{l}{$r_{co_{12}}:  $}& \multicolumn{3}{l}{ $ \rightarrow  CE{t_0}90$ }\\
\multicolumn{3}{l}{$r_{fcg_{13}}: $}& \multicolumn{3}{l}{ $ \Rightarrow CS_g{t_0}90$}&% ~~~~~~~~~~ &
\multicolumn{3}{l}{$r_{fcg_{23}}: $}& \multicolumn{3}{l}{ $ \Rightarrow  CS_g{t_1}90$}& %~~~~~&
\multicolumn{3}{l}{$r_{fcg_{33}}: $}& \multicolumn{3}{l}{ $ \Rightarrow  CS_g{t_2}90$}&% ~~~~~&
\multicolumn{3}{l}{$r_{co_{13}}:  $}& \multicolumn{3}{l}{ $ \rightarrow  CS{t_0}90$}\\

\multicolumn{24}{l}{~}\\

\multicolumn{3}{l}{$r_{fce_{11}}: $}& \multicolumn{3}{l}{$ \Rightarrow CN_e{t_0}90$}&%~~~~~~~~~~ &
\multicolumn{3}{l}{$r_{fce_{21}}: $}& \multicolumn{3}{l}{$ \Rightarrow CN_e{t_1}75$}&%~~~~~&
\multicolumn{3}{l}{$r_{fce_{31}}: $}& \multicolumn{3}{l}{$ \Rightarrow CN_e{t_2}30$}&%\\
\multicolumn{3}{l}{~}& \multicolumn{3}{l}{~}\\
\multicolumn{3}{l}{$r_{fce_{12}}: $}& \multicolumn{3}{l}{$ \Rightarrow CC_e{t_0}90$}& %~~~~~~~~~~ &
\multicolumn{3}{l}{$r_{fce_{22}}: $}& \multicolumn{3}{l}{$ \Rightarrow  CC_e{t_1}75$}&%~~~~~&
\multicolumn{3}{l}{$r_{fce_{32}}: $}& \multicolumn{3}{l}{$ \Rightarrow  CC_e{t_2}30$}&%
\multicolumn{3}{l}{~}& \multicolumn{3}{l}{~}\\
\multicolumn{3}{l}{$r_{fce_{13}}: $}& \multicolumn{3}{l}{$ \Rightarrow CS_e{t_0}90$}& %~~~~~~~~~~ &
\multicolumn{3}{l}{$r_{fce_{23}}: $}& \multicolumn{3}{l}{$ \Rightarrow  CS_e{t_1}75$}&%~~~~~&
\multicolumn{3}{l}{$r_{fce_{33}}: $}& \multicolumn{3}{l}{$ \Rightarrow  CS_e{t_2}30$}&%
\multicolumn{3}{l}{~}& \multicolumn{3}{l}{~}\\

\multicolumn{24}{l}{~}\\

\multicolumn{3}{l}{$r_{wg_{11}}:$} & \multicolumn{3}{l}{$ \Rightarrow WN_g{t_0}N18$} &%
\multicolumn{3}{l}{$r_{wg_{21}}:$} & \multicolumn{3}{l}{$ \Rightarrow WN_g{t_1}N8$} &%~&
\multicolumn{3}{l}{$r_{wg_{31}}:$} & \multicolumn{3}{l}{$ \Rightarrow  WN_g{t_2}N8$} &%~&
\multicolumn{3}{l}{$r_{wo_{11}}:$} & \multicolumn{3}{l}{$ \rightarrow  WN{t_0}NE15$}\\
\multicolumn{3}{l}{$r_{wg_{12}}:$} & \multicolumn{3}{l}{$ \Rightarrow WC_g{t_0}N18$} &%~&
\multicolumn{3}{l}{$r_{wg_{22}}:$} & \multicolumn{3}{l}{$ \Rightarrow WC_g{t_1}E8$}&%~&
\multicolumn{3}{l}{$r_{wg_{32}}:$} & \multicolumn{3}{l}{$ \Rightarrow WC_g{t_2}E8$}&%~&
\multicolumn{3}{l}{$r_{wo_{12}}:$} & \multicolumn{3}{l}{$ \rightarrow WC{t_0}NE15$} \\
\multicolumn{3}{l}{$r_{wg_{13}}:$} & \multicolumn{3}{l}{$ \Rightarrow WS_g{t_0}N10$} &%~&
\multicolumn{3}{l}{$r_{wg_{23}}:$} & \multicolumn{3}{l}{$ \Rightarrow WS_g{t_1}E5$} &%~&
\multicolumn{3}{l}{$r_{wg_{33}}:$} & \multicolumn{3}{l}{$ \Rightarrow WS_g{t_2}E5$} &%~&
\multicolumn{3}{l}{$r_{wo_{13}}:$} & \multicolumn{3}{l}{$ \rightarrow WS{t_0}NE15$}\\

\multicolumn{24}{l}{~}\\

\multicolumn{3}{l}{$r_{we_{11}}:$} & \multicolumn{3}{l}{$ \Rightarrow WN_e{t_1}NE15$} &%~&
\multicolumn{3}{l}{$r_{we_{21}}:$} & \multicolumn{3}{l}{$ \Rightarrow WN_e{t_1}NE5$}&%~&
\multicolumn{3}{l}{$r_{we_{31}}:$} & \multicolumn{3}{l}{$ \Rightarrow  WN_e{t_2}N5$}&%
\multicolumn{3}{l}{~}& \multicolumn{3}{l}{~}\\
\multicolumn{3}{l}{$r_{we_{12}}:$} & \multicolumn{3}{l}{$ \Rightarrow WC_e{t_1}NE5$} &%~&
\multicolumn{3}{l}{$r_{we_{22}}:$} & \multicolumn{3}{l}{$ \Rightarrow  WC_e{t_1}NE5$}&%~&
\multicolumn{3}{l}{$r_{we_{32}}:$} & \multicolumn{3}{l}{$ \Rightarrow  WC_e{t_2}N5$}&%
\multicolumn{3}{l}{~}& \multicolumn{3}{l}{~}\\
\multicolumn{3}{l}{$r_{we_{13}}:$} & \multicolumn{3}{l}{$ \Rightarrow WS_e{t_1}N5$} &%~&
\multicolumn{3}{l}{$r_{we_{23}}:$} & \multicolumn{3}{l}{$ \Rightarrow  WS_e{t_1}NE5$}&%~&
\multicolumn{3}{l}{$r_{we_{33}}:$} & \multicolumn{3}{l}{$ \Rightarrow  WS_e{t_2}N5$}&%
\multicolumn{3}{l}{~}& \multicolumn{3}{l}{~}\\

\multicolumn{24}{l}{~}\\

\multicolumn{3}{l}{$r_{sg_{11}}:$} & \multicolumn{3}{l}{$ \Rightarrow Sea_g{t_0}190$} &%~&
\multicolumn{3}{l}{$r_{sg_{21}}:$} & \multicolumn{3}{l}{$ \Rightarrow  Sea_g{t_1}100$} &%~&
\multicolumn{3}{l}{$r_{sg_{31}}:$} & \multicolumn{3}{l}{$ \Rightarrow  Sea_g{t_2}100$} &%~&
\multicolumn{3}{l}{$r_{so_{11}}:$} & \multicolumn{3}{l}{$ \rightarrow  Sea_o{t_0}190$}\\
\multicolumn{3}{l}{$r_{se_{11}}:$} & \multicolumn{3}{l}{$ \Rightarrow Sea_e{t_0}160$} &%~&
\multicolumn{3}{l}{$r_{se_{21}}:$} & \multicolumn{3}{l}{$ \Rightarrow  Sea_e{t_1}50$}&%~&  
\multicolumn{3}{l}{$r_{se_{31}}:$} & \multicolumn{3}{l}{$ \Rightarrow  Sea_e{t_2}10$}   &%~& &
\multicolumn{3}{l}{~}& \multicolumn{3}{l}{~}\\

\multicolumn{24}{l}{~}\\
\hline
\multicolumn{24}{l}{~}\\

\multicolumn{3}{l}{$r_{cg_{11}}:$} & \multicolumn{9}{l}{$ CN_g{t_1}90, CN_e{t_1}75 \Rightarrow CN{t_1}88$} &%~~~ &
\multicolumn{3}{l}{$r_{ce_{11}}:$} & \multicolumn{9}{l}{$  CN_g{t_1}90, CN_e{t_1}75 \Rightarrow CN{t_1}78$}\\
\multicolumn{3}{l}{$r_{cg_{12}}:$} & \multicolumn{9}{l}{$ CC_g{t_1}90, CC_e{t_1}75 \Rightarrow CC{t_1}88$} &%~~~ &
\multicolumn{3}{l}{$r_{ce_{12}}:$} & \multicolumn{9}{l}{$  CC_g{t_1}90, CC_e{t_1}75 \Rightarrow CC{t_1}78$}\\
\multicolumn{3}{l}{$r_{cg_{13}}:$} & \multicolumn{9}{l}{$ CS_g{t_1}90, CS_e{t_1}75 \Rightarrow CS{t_1}88$} &%~~~ &
\multicolumn{3}{l}{$r_{ce_{13}}:$} & \multicolumn{9}{l}{$  CS_g{t_1}90, CS_e{t_1}75 \Rightarrow CS{t_1}78$}\\
\multicolumn{3}{l}{$r_{wg_{11}}:$} & \multicolumn{9}{l}{$ WN_g{t_1}N8, WN_e{t_1}NE5 \Rightarrow WN{t_1}N7$} &%~~~ &
\multicolumn{3}{l}{$r_{we_{11}}:$} & \multicolumn{9}{l}{$  WN_g{t_1}N8, WN_e{t_1}NE5 \Rightarrow WN{t_1}NE6$}\\
\multicolumn{3}{l}{$r_{wg_{12}}:$} & \multicolumn{9}{l}{$ WC_g{t_1}E8, WC_e{t_1}NE5 \Rightarrow WC{t_1}E7$} &%~~~ &
\multicolumn{3}{l}{$r_{we_{12}}:$} & \multicolumn{9}{l}{$  WC_g{t_1}E8, WC_e{t_1}NE5 \Rightarrow WC{t_1}NE6$}\\
\multicolumn{3}{l}{$r_{wg_{13}}:$} & \multicolumn{9}{l}{$ WS_g{t_1}E5, WS_e{t_1}N5 \Rightarrow WS{t_1}E5$} &%~~~&
\multicolumn{3}{l}{$r_{we_{13}}:$} & \multicolumn{9}{l}{$ WS_g{t_1}E5, WS_e{t_1}N5 \Rightarrow WS{t_1}N5$}\\
\multicolumn{3}{l}{$r_{sg_{11}}:$} & \multicolumn{9}{l}{$ Sea_g{t_1}100, Sea_e{t_1}50 \Rightarrow Sea_{t_1}95$} &%~~~&
\multicolumn{3}{l}{$r_{se_{11}}:$} & \multicolumn{9}{l}{$ Sea_g{t_1}100, Sea_e{t_1}50 \Rightarrow Sea_{t_1}65$}\\

\multicolumn{24}{l}{~}\\

\multicolumn{3}{l}{$v_{c_{11}}:$} & \multicolumn{5}{l}{$CN{t_1}88  \Rightarrow \neg CN{t_1}78 $} &%~&
\multicolumn{3}{l}{$v_{c_{12}}:$} & \multicolumn{5}{l}{$CC{t_1}88  \Rightarrow \neg CC{t_1}78 $} &%
\multicolumn{3}{l}{$v_{c_{11}}:$} & \multicolumn{5}{l}{$CS{t_1}88  \Rightarrow \neg CS{t_1}78 $} \\
\multicolumn{3}{l}{$v_{c_{21}}:$} & \multicolumn{5}{l}{$CN{t_1}78  \Rightarrow \neg CN{t_1}88 $} &%~&
\multicolumn{3}{l}{$v_{c_{22}}:$} & \multicolumn{5}{l}{$CC{t_1}78  \Rightarrow \neg CC{t_1}88 $} &%
\multicolumn{3}{l}{$v_{c_{23}}:$} & \multicolumn{5}{l}{$CS{t_1}78  \Rightarrow \neg CS{t_1}78 $}\\
\multicolumn{3}{l}{$v_{w_{11}}:$} & \multicolumn{5}{l}{$WN{t_1}N7  \Rightarrow \neg WN{t_1}NE6 $} &%~&
\multicolumn{3}{l}{$v_{w_{12}}:$} & \multicolumn{5}{l}{$WC{t_1}E7  \Rightarrow \neg WC{t_1}NE6 $} &%
\multicolumn{3}{l}{$v_{w_{11}}:$} & \multicolumn{5}{l}{$WS{t_1}E5  \Rightarrow \neg WS{t_1}NE5 $}\\
\multicolumn{3}{l}{$v_{w_{21}}:$} & \multicolumn{5}{l}{$WN{t_1}NE6  \Rightarrow \neg WN{t_1}N7 $} &%~&
\multicolumn{3}{l}{$v_{w_{22}}:$} & \multicolumn{5}{l}{$WC{t_1}NE6  \Rightarrow \neg WC{t_1}E7 $} &%
\multicolumn{3}{l}{$v_{w_{23}}:$} & \multicolumn{5}{l}{$WS{t_1}NE5  \Rightarrow \neg WS{t_1}E5 $}\\
%
%\multicolumn{24}{l}{~}\\
\multicolumn{3}{l}{$v_{s_{11}}:$} & \multicolumn{5}{l}{$Sea_{t_1}95  \Rightarrow \neg Sea_{t_2}75 $} &%~&~&~&
\multicolumn{3}{l}{$v_{s_{21}}:$} & \multicolumn{5}{l}{$Sea_{t_1}75  \Rightarrow \neg Sea_{t_2}95$}\\
\multicolumn{3}{l}{~}& \multicolumn{5}{l}{~}\\

\multicolumn{24}{l}{~}\\

\multicolumn{3}{l}{$r_{cg_{21}}:$} & \multicolumn{9}{l}{$ CN_g{t_2}90, CN_e{t_2}30 \Rightarrow CN{t_2}68$} &%~~~ &
\multicolumn{3}{l}{$r_{ce_{21}}:$}& \multicolumn{9}{l}{$  CN_g{t_2}90, CN_e{t_2}30 \Rightarrow CN{t_2}38$}\\
\multicolumn{3}{l}{$r_{cg_{22}}:$} & \multicolumn{9}{l}{$ CC_g{t_2}90, CC_e{t_2}30 \Rightarrow CC{t_2}68$} &%~~~ &
\multicolumn{3}{l}{$r_{ce_{22}}:$} & \multicolumn{9}{l}{$  CC_g{t_2}90, CC_e{t_2}30 \Rightarrow CC{t_2}38$}\\
\multicolumn{3}{l}{$r_{cg_{23}}:$} & \multicolumn{9}{l}{$ CE_g{t_2}90, CE_e{t_2}30 \Rightarrow CE{t_2}68$} &%~~~ &
\multicolumn{3}{l}{$r_{ce_{23}}:$} & \multicolumn{9}{l}{$  CE_g{t_2}90, CE_e{t_2}30 \Rightarrow CE{t_2}38$}\\
\multicolumn{3}{l}{$r_{wg_{21}}:$} & \multicolumn{9}{l}{$ WN_g{t_2}N8, WN_e{t_2}N5 \Rightarrow WN{t_2}NE7$} &%~~~ &
\multicolumn{3}{l}{$r_{we_{21}}:$}& \multicolumn{9}{l}{$  WN_g{t_2}N8, WN_e{t_2}N5 \Rightarrow WN{t_2}N6$}\\
\multicolumn{3}{l}{$r_{wg_{22}}:$} & \multicolumn{9}{l}{$ WC_g{t_2}E8, WC_e{t_2}N5 \Rightarrow WC{t_2}NE7$} &%~~~ &
\multicolumn{3}{l}{$r_{we_{22}}:$} & \multicolumn{9}{l}{$  WC_g{t_2}E8, WC_e{t_2}N5 \Rightarrow WC{t_2}N6$}\\
\multicolumn{3}{l}{$r_{wg_{23}}:$} & \multicolumn{9}{l}{$ WS_g{t_2}E5, WS_e{t_2}N5 \Rightarrow WS{t_2}NE5$} &%~~~&
\multicolumn{3}{l}{$r_{we_{23}}:$} & \multicolumn{9}{l}{$  WS_g{t_2}E5, WS_e{t_2}N5 \Rightarrow WS{t_2}N5$}\\
\multicolumn{3}{l}{$r_{sg_{21}}:$} & \multicolumn{9}{l}{$ Sea_g{t_2}100, Sea_e{t_2}10 \Rightarrow Sea_{t_2}80$} &%~~~&
\multicolumn{3}{l}{$r_{se_{21}}:$} & \multicolumn{9}{l}{$  Sea_g{t_2}100, Sea_e{t_2}10 \Rightarrow Sea_{t_2}20$}\\

\multicolumn{24}{l}{~}\\

\multicolumn{3}{l}{$v_{c_{31}}:$} & \multicolumn{5}{l}{$CN{t_2}68  \Rightarrow \neg CN{t_2}38 $} &%~&
\multicolumn{3}{l}{$v_{c_{32}}:$} & \multicolumn{5}{l}{$CC{t_2}68  \Rightarrow \neg CC{t_2}38 $} &%~~
\multicolumn{3}{l}{$v_{c_{31}}:$} & \multicolumn{5}{l}{$CS{t_2}68  \Rightarrow \neg CS{t_2}38 $}\\
\multicolumn{3}{l}{$v_{c_{41}}:$} & \multicolumn{5}{l}{$CN{t_2}38  \Rightarrow \neg CN{t_2}68 $ }&%~&
\multicolumn{3}{l}{$v_{c_{42}}:$} & \multicolumn{5}{l}{$CC{t_2}38  \Rightarrow \neg CC{t_2}68 $ }&%~~
\multicolumn{3}{l}{$v_{c_{43}}:$} & \multicolumn{5}{l}{$CS{t_2}38  \Rightarrow \neg CS{t_2}68 $}\\
\multicolumn{3}{l}{$v_{w_{31}}:$} & \multicolumn{5}{l}{$WN{t_2}NE7  \Rightarrow \neg WN{t_2}N6 $} &%~&
\multicolumn{3}{l}{$v_{w_{32}}:$} & \multicolumn{5}{l}{$WC{t_2}NE7  \Rightarrow \neg WC{t_2}N6 $} &%~~
\multicolumn{3}{l}{$v_{w_{31}}:$} & \multicolumn{5}{l}{$WS{t_2}NE5  \Rightarrow \neg WS{t_2}N5 $}\\
\multicolumn{3}{l}{$v_{w_{41}}:$} & \multicolumn{5}{l}{$WN{t_2}N6  \Rightarrow \neg WN{t_2}NE7 $} &%~&
\multicolumn{3}{l}{$v_{w_{42}}:$} & \multicolumn{5}{l}{$WC{t_2}N6  \Rightarrow \neg WC{t_2}NE7 $} &%~~
\multicolumn{3}{l}{$v_{w_{43}}:$} & \multicolumn{5}{l}{$WS{t_2}N5  \Rightarrow \neg WS{t_2}NE5 $}\\

\multicolumn{3}{l}{$v_{s_{31}}:$} & \multicolumn{5}{l}{$ Sea_{t_2}80  \Rightarrow \neg Sea_{t_2}20 $} &%~&~&~&
\multicolumn{3}{l}{$v_{s_{41}}:$} & \multicolumn{5}{l}{$Sea_{t_2}20  \Rightarrow \neg Sea_{t_2}80 $} &
\multicolumn{3}{l}{~}& \multicolumn{5}{l}{~}\\

\multicolumn{24}{l}{~}\\
\hline
\multicolumn{24}{l}{~}\\

\multicolumn{3}{l}{$p_{11}: $} & \multicolumn{3}{l}{$ v_{c_{21}}   ~ \rangle ~  r_{cg_{11}} $}&%~~~~~~ &
\multicolumn{3}{l}{$p_{12}: $} & \multicolumn{3}{l}{$ r_{ce_{11}} ~ \rangle ~  v_{c_{11}} $}&%~~~ &
\multicolumn{3}{l}{$p_{13}: $} & \multicolumn{3}{l}{$ v_{c_{41}}   ~ \rangle ~  r_{cg_{21}} $}&%~~~~~ &
\multicolumn{3}{l}{$p_{14}: $} & \multicolumn{3}{l}{$ r_{ce_{21}} ~ \rangle ~  v_{c_{31}} $}\\
\multicolumn{3}{l}{$p_{21}: $} & \multicolumn{3}{l}{$ v_{w_{21}}  ~ \rangle ~  r_{wg_{11}} $}&%~~~~    &
\multicolumn{3}{l}{$p_{22}: $} & \multicolumn{3}{l}{$ r_{we_{11}} ~ \rangle ~  v_{w_{11}} $}&%~~~ &
\multicolumn{3}{l}{$p_{23}: $} & \multicolumn{3}{l}{$ v_{w_{41}}   ~ \rangle ~  r_{wg_{21}} $}&%~~~~~&
\multicolumn{3}{l}{$p_{24}: $} & \multicolumn{3}{l}{$ r_{we_{21}} ~ \rangle ~  v_{w_{31}} $}\\

\multicolumn{3}{l}{$p_{31}: $} & \multicolumn{3}{l}{$ v_{c_{22}}   ~ \rangle ~  r_{cg_{12}} $}&%~~~~~~ &
\multicolumn{3}{l}{$p_{32}: $} & \multicolumn{3}{l}{$ r_{ce_{12}} ~ \rangle ~  v_{c_{12}} $} &%~~~ &
\multicolumn{3}{l}{$p_{33}: $} & \multicolumn{3}{l}{$ v_{c_{42}}   ~ \rangle ~  r_{cg_{22}} $}&%~~~~~ &
\multicolumn{3}{l}{$p_{34}: $} & \multicolumn{3}{l}{$ r_{ce_{22}} ~ \rangle ~  v_{c_{32}} $}\\
\multicolumn{3}{l}{$p_{41}: $} & \multicolumn{3}{l}{$ v_{w_{22}}  ~ \rangle ~  r_{wg_{12}} $}&%~~~~    &
\multicolumn{3}{l}{$p_{42}: $} & \multicolumn{3}{l}{$ r_{we_{12}} ~ \rangle ~  v_{w_{12}} $} &%~~~ &
\multicolumn{3}{l}{$p_{43}: $} & \multicolumn{3}{l}{$ v_{w_{42}}   ~ \rangle ~  r_{wg_{22}} $}&%~~~~~&
\multicolumn{3}{l}{$p_{44}: $} & \multicolumn{3}{l}{$ r_{we_{22}} ~ \rangle ~  v_{w_{32}} $}\\

\multicolumn{3}{l}{$p_{51}: $} & \multicolumn{3}{l}{$ v_{c_{23}}   ~ \rangle ~  r_{cg_{13}} $}&%~~~~~~ &
\multicolumn{3}{l}{$p_{52}: $} & \multicolumn{3}{l}{$ r_{ce_{13}} ~ \rangle ~  v_{c_{13}} $} &%~~~ &
\multicolumn{3}{l}{$p_{53}: $} & \multicolumn{3}{l}{$ v_{c_{43}}   ~ \rangle ~  r_{cg_{23}} $}&%~~~~~ &
\multicolumn{3}{l}{$p_{54}: $} & \multicolumn{3}{l}{$ r_{ce_{23}} ~ \rangle ~  v_{c_{33}} $} \\
\multicolumn{3}{l}{$p_{61}: $} & \multicolumn{3}{l}{$ v_{w_{23}}  ~ \rangle ~  r_{wg_{13}} $}&%~~~~    &
\multicolumn{3}{l}{$p_{62}: $} & \multicolumn{3}{l}{$ r_{we_{13}} ~ \rangle ~  v_{w_{13}} $} &%~~~ &
\multicolumn{3}{l}{$p_{63}: $} & \multicolumn{3}{l}{$ v_{w_{43}}   ~ \rangle ~  r_{wg_{23}} $}&%~~~~~&
\multicolumn{3}{l}{$p_{64}: $} & \multicolumn{3}{l}{$ r_{we_{23}} ~ \rangle ~  v_{w_{33}} $}\\

\multicolumn{3}{l}{$p_{71}: $} & \multicolumn{3}{l}{$ v_{s_{21}}   ~ \rangle ~  r_{sg_{11}} $}&%~~~~~~ &
\multicolumn{3}{l}{$p_{72}: $} & \multicolumn{3}{l}{$ r_{se_{11}} ~ \rangle ~  v_{s_{11}} $} &%~~~ &
\multicolumn{3}{l}{$p_{73}: $} & \multicolumn{3}{l}{$ v_{s_{41}}   ~ \rangle ~  r_{sg_{21}} $}&%~~~~~ &
\multicolumn{3}{l}{$p_{74}: $} & \multicolumn{3}{l}{$ r_{se_{21}} ~ \rangle ~  v_{s_{31}} $}
\end{tabular} 
}
\end{center}

Given this theory, the \emph{Reasoner} concludes $+\partial CN{t_1}78$, $+\partial CC{t_1}78$, $+\partial CS{t_1}78$, $+\partial WN{t_1}NE6$, $+\partial WC{t_1}NE6$, $+\partial WS{t_1}N5$, $+\partial Sea_{t_1}65$,$+\partial CN{t_2}38$, $+\partial CC{t_2}38$, $+\partial CS{t_2}38$, $+\partial WN{t_2}N6$, $+\partial WC{t_2}N6$, $+\partial WS{t_2}N5$, $+\partial Sea_{t_2}20$. %\footnote{The full implementation with the SPINdle engine  is available online at \ARXIVSPINDLE}

\begin{figure}[!h]
\centering
\includegraphics[width=3.8cm]{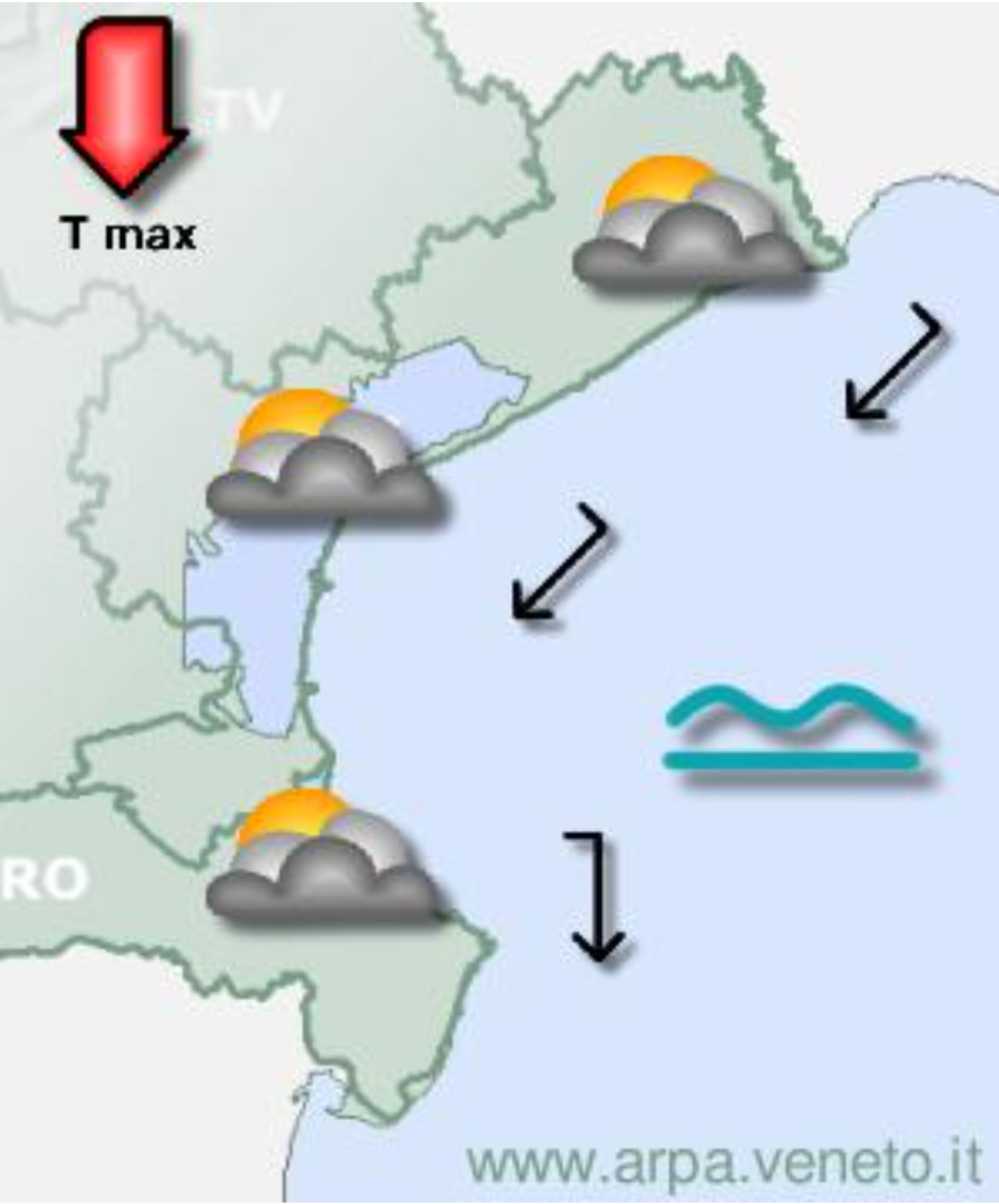}   \includegraphics[width=3.8cm]{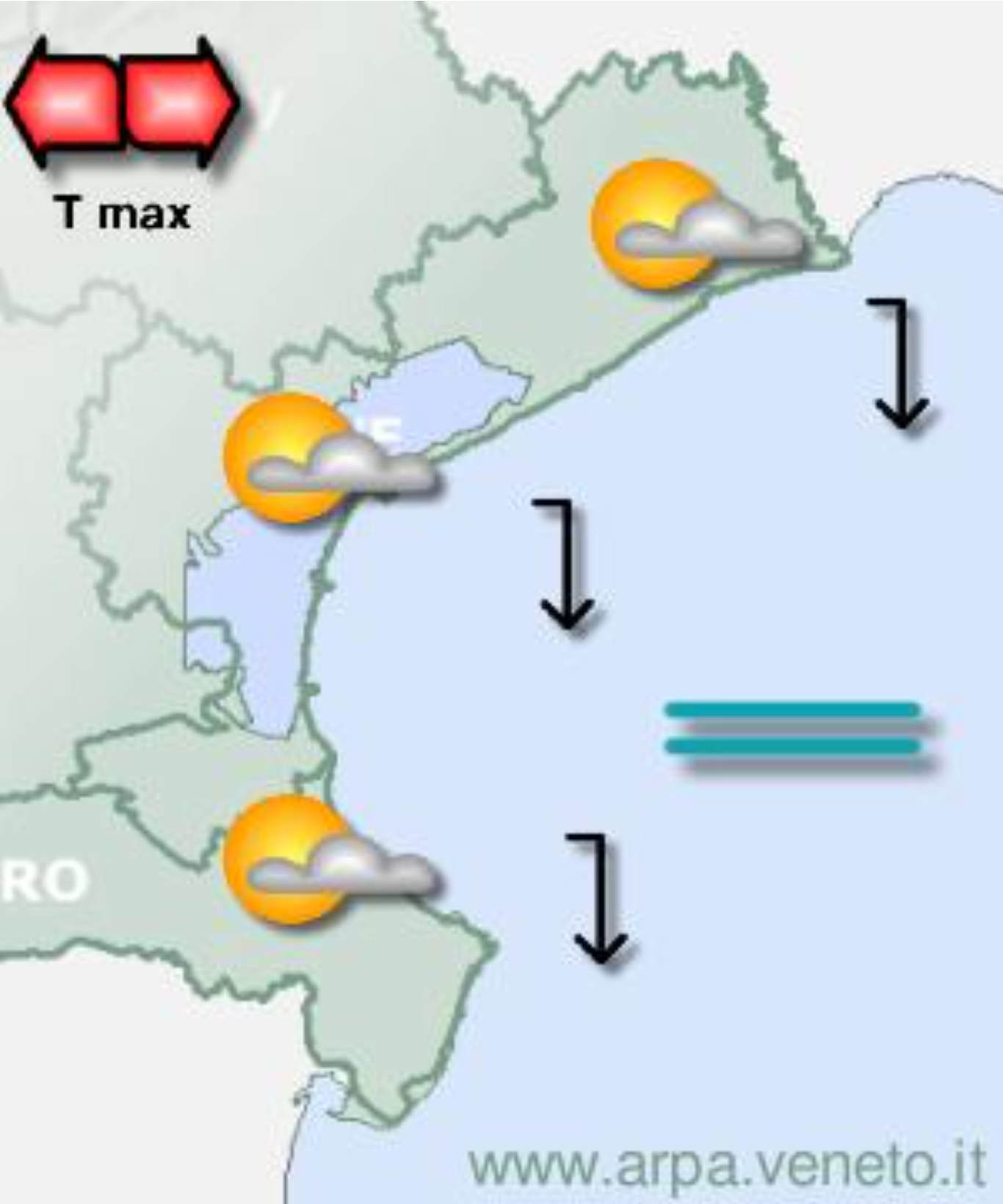} 
%at $t_1$ (tomorrow)&at $t_2$  (day after  $t_1$)\\ 

\caption{Weather forecast at $t_1$ (tomorrow; left and at $t_2$ (day after tomorrow; right)}
\label{forecastExample}
\end{figure}

Therefore, translating numerical value into words, we have at at time $t_1$
\begin{center}
{\small
\begin{tabular}{lcl} North &:& Mostly Cloudy, Light Winds from North East \\ Center &:& Mostly Cloudy, Light Winds from North East \\South&:&Mostly Cloudy, Light Winds from North \\Sea&: &  Slight
\end{tabular}
}
\end{center}
and at time $t_2$   
\begin{center}
{\small
\begin{tabular}{lcl} North &:& Partly  Cloudy, Light Winds from North \\ Center &:& Partly  Cloudy, Light Winds from North \\South&:&Partly  Cloudy, Light Winds from North \\Sea&: &  Calm 
\end{tabular}
}
\end{center}
which can be expressed iconographically as in Figure \ref{forecastExample}.% (courtesy of Arpa Veneto, Italy).\fix{Luca}{Io toglierei questo ``courtesy'' e invece chiarire il contributo di ARPAV (e scriverei ARPAV o Arpa Veneto, non tutti e due)}

%% file: SPINDLE_CONCLUSIONS.tex
\section{SPINdle}
 One the \HL\ algorithm described in the paper produced a defeasible theory, we can process the theory by means of well-established reasoning technologies, such as Spindle. SPINdle is a logic reasoner that can be used to compute the consequence of defeasible logic theories in an efficient and it can be downloaded at http://spindle.data61.csiro.au/spindle/.
\subsection{Spindle conclusions for rules of the reference implementation}\label{sec:spindle1}
{\small 
\begin{verbatim}
*****************************************************************************
* SPINdle (version 2.2.4)                                                   *
* Copyright (C) 2009-2014 NICTA Ltd.                                        *
* This software and its documentation is distributed under the terms of the *
* FSF Lesser GNU Public License (LGPL).                                     *
*                                                                           *
* This program comes with ABSOLUTELY NO WARRANTY; This is a free software   *
* and you are welcome to redistribute it under certain conditions; for      *
* details type:                                                             *
* java -jar spindle-<version>.jar --app.license                             *
*****************************************************************************
=========================
== application start!! ==
=========================
Initialize application context - start
    load application configuration - start
        app.showProgress=false
        app.showStatistics=false
        reasoner.version=2
    load application configuration - end
    configurating I/O classes - start
        generating outputter [spindle.io.outputter.DflTheoryOutputter]...success, type=[dfl]
        generating outputter [spindle.io.outputter.XmlTheoryOutputter2]...success, type=[xml]
        generating parser [spindle.io.parser.DflTheoryParser2]...success, type=[dfl]
        generating parser [spindle.io.parser.XmlTheoryParser2]...success, type=[xml]
    configurating I/O classes - end
Initialize application context - end
=== System info: Load theory from url: file:/temp/Meteo_SPINDLE_RULES
=== System info: Theory loaded successfully, theory type: SDL.
=== System info: Theory contains no literal variable or boolean function.
=== System info: transform theory to regular form
=== System info: Generate conclusions.
=== System info: Conclusions.
    +D CEt090(X)
    +D CNt090(X)
    +D CSt090(X)
    +D Seaot0190(X)
    +D WCt0NE15(X)
    +D WNt0NE15(X)
    +D WSt0NE15(X)
    +D Wo_t0_5(X)
    -D CCet175(X)
    -D CCet190(X)
    -D CCet230(X)
    -D CCgt090(X)
    -D CCgt190(X)
    -D CCgt290(X)
    -D CCt178(X)
    -D -CCt178(X)
    -D CCt188(X)
    -D -CCt188(X)
    -D CCt238(X)
    -D -CCt238(X)
    -D CCt268(X)
    -D -CCt268(X)
    -D CNet175(X)
    -D CNet190(X)
    -D CNet230(X)
    -D CNgt090(X)
    -D CNgt190(X)
    -D CNgt290(X)
    -D CNt178(X)
    -D -CNt178(X)
    -D CNt188(X)
    -D -CNt188(X)
    -D CNt238(X)
    -D -CNt238(X)
    -D CNt268(X)
    -D -CNt268(X)
    -D CSet175(X)
    -D CSet190(X)
    -D CSet230(X)
    -D CSgt090(X)
    -D CSgt190(X)
    -D CSgt290(X)
    -D CSt178(X)
    -D -CSt178(X)
    -D CSt188(X)
    -D -CSt188(X)
    -D CSt238(X)
    -D -CSt238(X)
    -D CSt268(X)
    -D -CSt268(X)
    -D Seaet0160(X)
    -D Seaet150(X)
    -D Seaet210(X)
    -D Seagt0190(X)
    -D Seagt1100(X)
    -D Seagt2100(X)
    -D Seat165(X)
    -D -Seat165(X)
    -D Seat195(X)
    -D -Seat195(X)
    -D Seat220(X)
    -D -Seat220(X)
    -D Seat280(X)
    -D -Seat280(X)
    -D WCet1NE5(X)
    -D WCet2N5(X)
    -D WCgt0N18(X)
    -D WCgt1E8(X)
    -D WCgt2E8(X)
    -D WCt1E7(X)
    -D -WCt1E7(X)
    -D WCt1NE6(X)
    -D -WCt1NE6(X)
    -D WCt2N6(X)
    -D -WCt2N6(X)
    -D WCt2NE7(X)
    -D -WCt2NE7(X)
    -D WNet1NE15(X)
    -D WNet1NE5(X)
    -D WNet2N5(X)
    -D WNgt0N18(X)
    -D WNgt1N8(X)
    -D WNgt2N8(X)
    -D WNt1N7(X)
    -D -WNt1N7(X)
    -D WNt1NE6(X)
    -D -WNt1NE6(X)
    -D WNt2N6(X)
    -D -WNt2N6(X)
    -D WNt2NE7(X)
    -D -WNt2NE7(X)
    -D WSet1N5(X)
    -D WSet1NE5(X)
    -D WSet2N5(X)
    -D WSgt0N10(X)
    -D WSgt1E5(X)
    -D WSgt2E5(X)
    -D WSt1E5(X)
    -D -WSt1E5(X)
    -D WSt1N5(X)
    -D -WSt1N5(X)
    -D WSt2N5(X)
    -D -WSt2N5(X)
    -D WSt2NE5(X)
    -D -WSt2NE5(X)
    +d CCet175(X)
    +d CCet190(X)
    +d CCet230(X)
    +d CCgt090(X)
    +d CCgt190(X)
    +d CCgt290(X)
    +d CCt178(X)
    +d -CCt188(X)
    +d CCt238(X)
    +d -CCt268(X)
    +d CEt090(X)
    +d CNet175(X)
    +d CNet190(X)
    +d CNet230(X)
    +d CNgt090(X)
    +d CNgt190(X)
    +d CNt090(X)
    +d CNt178(X)
    +d -CNt188(X)
    +d CSet175(X)
    +d CSet190(X)
    +d CSet230(X)
    +d CSgt090(X)
    +d CSgt190(X)
    +d CSgt290(X)
    +d CSt090(X)
    +d CSt178(X)
    +d -CSt188(X)
    +d CSt238(X)
    +d -CSt268(X)
    +d Seaet0160(X)
    +d Seaet150(X)
    +d Seaet210(X)
    +d Seagt0190(X)
    +d Seagt1100(X)
    +d Seagt2100(X)
    +d Seaot0190(X)
    +d Seat165(X)
    +d -Seat195(X)
    +d Seat220(X)
    +d -Seat280(X)
    +d WCet1NE5(X)
    +d WCet2N5(X)
    +d WCgt0N18(X)
    +d WCgt1E8(X)
    +d WCgt2E8(X)
    +d WCt0NE15(X)
    +d -WCt1E7(X)
    +d WCt1NE6(X)
    +d WCt2N6(X)
    +d -WCt2NE7(X)
    +d WNet1NE15(X)
    +d WNet1NE5(X)
    +d WNet2N5(X)
    +d WNgt0N18(X)
    +d WNgt1N8(X)
    +d WNgt2N8(X)
    +d WNt0NE15(X)
    +d -WNt1N7(X)
    +d WNt1NE6(X)
    +d WNt2N6(X)
    +d -WNt2NE7(X)
    +d WSet1N5(X)
    +d WSet1NE5(X)
    +d WSet2N5(X)
    +d WSgt0N10(X)
    +d WSgt1E5(X)
    +d WSgt2E5(X)
    +d WSt0NE15(X)
    +d -WSt1E5(X)
    +d WSt1N5(X)
    +d WSt2N5(X)
    +d -WSt2NE5(X)
    +d Wo_t0_5(X)
    -d -CCt178(X)
    -d CCt188(X)
    -d -CCt238(X)
    -d CCt268(X)
    -d CNgt290(X)
    -d -CNt178(X)
    -d CNt188(X)
    -d CNt238(X)
    -d -CNt238(X)
    -d CNt268(X)
    -d -CNt268(X)
    -d -CSt178(X)
    -d CSt188(X)
    -d -CSt238(X)
    -d CSt268(X)
    -d -Seat165(X)
    -d Seat195(X)
    -d -Seat220(X)
    -d Seat280(X)
    -d WCt1E7(X)
    -d -WCt1NE6(X)
    -d -WCt2N6(X)
    -d WCt2NE7(X)
    -d WNt1N7(X)
    -d -WNt1NE6(X)
    -d -WNt2N6(X)
    -d WNt2NE7(X)
    -d WSt1E5(X)
    -d -WSt1N5(X)
    -d -WSt2N5(X)
    -d WSt2NE5(X)

====================================
== Performance statistics summary ==
====================================
== I/O classes configuration time used: 32 ms
== No. of record(s) found: 1
== --- start
\end{verbatim}
}
{\tiny
\begin{verbatim}
+------------+------------+-----------------+-----------------+-----------------+-----------------+-----------------+-------------+-----
|   No. of   |   No. of   |   Time used on  |   Time used on  |   Time used on  |  Time used on   |    Total time   | Max. Memory |
|    Rules   |  Literals  |  loading theory | transform theory| remove defeater |    reasoning    |       used      |     used    | filename
+------------+------------+-----------------+-----------------+-----------------+-----------------+-----------------+-------------+-----
|        105 |        105 |       0,069 sec |       0,006 sec |       0,000 sec |       0,035 sec |       0,110 sec |     9,63 MB | file:/temp/Meteo_SPINDLE_RULES
+------------+------------+-----------------+-----------------+-----------------+-----------------+-----------------+-------------+-----
== --- end
\end{verbatim}
}
{\small
\begin{verbatim}
Calling the shutdown routine...
Terminate application context - start
Terminate application context - end
=======================================
=== Application shutdown completed! ===
=======================================
\end{verbatim}
}

%% file: related.tex
\label{sec:related}
%The usage of intelligent decision technologies in the weather forecast process is a long-term research effort. 
Since the pioneering studies~\cite{Conway198923,Desmarais1990150,McCarthy1990228} and further engineering investigations on the commercial solutions~\cite{Moninger1990457}, a first attempt going in the same direction that we following in this paper appeared in the 1990s~\cite{Goldberg1993156} and inspired many specialized studies %thenceforth, in many different specialised fields including snow research, marine forecasting and agrimeteorology 
\cite{Kumar1994373,Marra1996475,Hansen199759,Fohn199819,CarrIII2001355,Mahabir20033749,Domanska20127673,Domanska201419}. The ontological approach and the usage of the Internet of Things have been applied to forecasting quite recently~\cite{Agresta2014417,Kulkarni2018555} and we acknowledge that the main technical inspirations of our framework trace back these works, whereas the main influences come from
 the usage of non-monotonic deduction systems for sensor-based applications (clearly related to the initial part of the forecasting process)~\cite{Tomazzoli2017345,claudio16}, and
non-monotonic reasoning\cite{Governatori2016296,Olivieri2013213,Governatori2016230,Governatori2014168}.%\fix{Luca}{Non capisco questa ultima frase. Manca il verbo e non capisco la parola ``topics''. Penso sia necessario scrivere in maniera pi\`u semplice e meno ``italiana''. Mi sembra anche ci sia un'esagerazione di autocitazioni. La met\`a delle references basterebbe e avanzerebbe, e direi di ridurre ad un terzo le autocitazioni di papers di cui siamo coautori. Ad esempio, \`e inutile citare il paper 29 se citiamo il 30, e il paper 21 se citiamo il 22, ed \`e inutile citare 10 papers di Governatori et al. Tanto queste citazioni non contano (o non dovrebbero contare) per l'h-index. Si pu\`o anche risparmiare molto spazio eliminando le informazioni non necessario nelle citazioni, ad esempio nella 2 la parentesi ``(including...)'' non serve a nulla} %A specific effort on the usage of non-monotonic deduction systems in sensor-based applications (clearly related to the initial part of the forecasting process) has been carried out by some of the authors in the recent past~\cite{Tomazzoli2017345,claudio16}.

%% file: conclusions.tex
In this paper, we proposes an architecture to support meteoroligists in  producing weather forecasts. The basic work is a reasoning framework able to simulate in a quite refined way the decision process made by the forecasters in producing weather bulletins.
There are several ways of extending this study. 
The research team includes a forecaster of the \arpav \ weather forecasting service, one of the most valuable forecasting service in Italy, who will lead the development of both the definition of the \emph{supremacy} function and the \HL\ algorithm.

%The very same architecture could be significantly extended. 
%We claim\fix{Luca}{In che senso claim?} that we design a portable approach, %One basic issue regards the portability of the
%approach,
%that has been intentionally devised for situations quite different from the Veneto area, where the project we are working at is in place.\fix{Luca}{Altro riferimento pericoloso, visto che non abbiamo mai parlato di questo nel paper. Questo va chiarito assolutamente (e in maniera diretta e semplice). Inoltre non si capisce in che senso questo sia un future work} 

We are currently working at the full formal definition the logical framework $\logic$. 
We plan to include more specific features in order to improve the precision of the automatic bulletin, aiming to a completely automatic and (potentially) unsupervised bulletin generator. 

%% file: tournament.tex
\subsubsection{The \HL\ algorithm}
\label{algorithm}

\begin{algorithm}[!h]
%\caption{\textbf{\nomeAlgo}:\\ Given an ordered set of \emph{Metarules}, a set of \emph{evaluations} and actual time, returns a defeasible theory}
\label{algo:METADEF}
{\scriptsize 
\begin{algorithmic}
\STATE  \textbf{Input:} a set of \emph{accuracy} of the methods $\Psi$, current time $\mathsf{t}$, an ordered set of \emph{Metarules} $\Theta$ in which metarules are in the form $label : assertion$ where labels are in the form $\langle method,time \rangle$ and $assertions$ are in the form $name(position,time,value)$;
\STATE \textbf{Output:} a defeasible theory $\cal{T}$ $= \langle F, R, P \rangle$ (facts, rules, priorities);
\STATE $\Theta^{\prime} \gets \Theta$
\REPEAT
	\STATE $m \gets pop(\Theta^{\prime})$; 
	\IF{$label(m).time \leq \mathsf{t} $}
		\IF{$ label(m).method = \mathsf{'O'} $}
			\STATE $f \gets assertion(m);\qquad$  $\quad \qquad  \;\, F \gets push(f,F)$;
		\ELSE	
			\STATE $r \gets \;\, \emptyset \Rightarrow assertion(m)$; $\quad R \gets push(r,R)$; \hspace{3mm} // \emph{r is in the form $A(r) \Rightarrow C(r)$ }
			\STATE  $name \gets C(r).name$; $position \gets C(r).position$; $time \gets C(r).time$;
			\STATE $\overline{R} \gets R - r $; 
			\REPEAT
				\STATE $\overline{r} \gets pop( \overline{R})$;   $\overline{l} \gets getLabel(\Theta, C(\overline{r}))$; 
				\STATE  $\overline{name} \gets C(\overline{r}).name$; $\overline{position} \gets C(\overline{r}).position$; $\overline{time} \gets C(\overline{r}).time$; 
				\IF{$name = \overline{name} $}
					\IF{$time = \overline{time} $}
						\IF{$position = \overline{position} $}
							\IF{$C(r).value \neq C(\overline{r}).value $}
								\STATE $sr_1 ~\quad \gets createNewRule()$; 
								\STATE $A(sr_1) \gets C(r) , C(\overline{r}) $;  $C(sr_1) \gets supremacy( C(r),C(\overline{r}),first )$;
								\STATE $sr_2 ~\quad \gets createNewRule()$;  
								\STATE $A(sr_2) \gets C(r) , C(\overline{r}) $; $C(sr_2) \gets supremacy( C(r),C(\overline{r}),last )$;
								\STATE $vc_1 ~\quad \gets createNewRule()$; 
								\STATE $A(vc_1) \gets C(r).value$; $C(vc_1) \gets \neg C(\overline{r}).value $;
								\STATE $vc_2 ~\quad \gets createNewRule()$; 
								\STATE $A(vc_2) \gets C(\overline{r}).value $;  $C(vc_2) \gets \neg C(r).value $;
								\STATE $accuracy_1 \gets get(\Psi, label(m).method)$;  $\quad accuracy_2 \gets get(\Psi,\overline{l}.method)$; 
								\IF{$accuracy_1 > accuracy_2$}
									\STATE $p_1 \gets crateNewPriority(sr_1,vr_2)$; $\qquad P \gets push(p_1, P)$;
									\STATE $p_2 \gets crateNewPriority(vr_1,sr_2)$; $\qquad P \gets push(p_2, P)$;
								\ELSIF {$accuracy_1 < accuracy_2$}
									\STATE $p_1 \gets crateNewPriority(sr_2,vr_1)$; $\qquad P \gets push(p_1, P)$;
									\STATE $p_2 \gets crateNewPriority(vr_2,sr_1)$; $\qquad P \gets push(p_2, P)$;
								\ELSE
									\IF{$label(m).time \geq \overline{l}.time$}
										\STATE $p_1 \gets crateNewPriority(sr_1,vr_2)$; $\quad P \gets push(p_1, P)$;
										\STATE $p_2 \gets crateNewPriority(vr_1,sr_2)$; $\quad P \gets push(p_2, P)$;
									\ELSE
										\STATE $p_1 \gets crateNewPriority(sr_2,vr_1)$; $\quad P \gets push(p_1, P)$;
										\STATE $p_2 \gets crateNewPriority(vr_2,sr_1)$; $\quad P \gets push(p_2, P)$;
									\ENDIF	
								\ENDIF
							\ENDIF
						\ENDIF
					\ENDIF
				\ENDIF			
			\UNTIL{ $\overline{R} = \emptyset $ };
		\ENDIF			
	\ENDIF
\UNTIL{ $\Theta^{\prime} = \emptyset $ };
\RETURN $\cal{T} = \langle$ F, R, P $ \rangle$ ;
\end{algorithmic}
}
\label{algo1}
\end{algorithm}
A certain rule is a candidate for rewriting, only if the synchroniser acknowledged that its clock time falls within the validity interval of the rule (if the rule has a validity interval) or at the exact instant of the rule if the rule is simply instantaneous.

%% file: referenceImpltwo.tex
\section{Another example}\label{refimpl2}
We would like make an example of a weather forecast considering our region, located in the north eastern part of Italy, to give a better evidence of how our model can fit a real life scenario. 
For the sake of space we will make some limitations: we limit the weather forecast to rain conditions and to only four points; we will label these points North, East ,South, West. We will use only two forecasting maps and we we will limit the time frame to only two values, representing two and one day after the present: respectively $t_2,t_1,t_0$

We have as input two forecasting sources, coming from different forecasting models such as IFS (also known as ECMWF for European Center Medium Weather Forecast)  and GFS (Global Forecast System), plus the map of observations.
The first source obtained with the GFS prevision model asserts, at time $t_0$  $\{North=5mm, East=4mm,South=4mm, West=4mm\}$, at time $t_1$  $\{North=4mm, East=4mm,South=4mm, West=5mm\}$, at time $t_2$  $\{North=6mm, East=6mm,South=6mm, West=6mm\}$. 
The second source obtained with the ECMWF prevision model asserts, at time $t_0$  $\{North=5mm, East=5mm,South=5mm, West=5mm\}$, at time $t_1$  $\{North=24mm, East=14mm,South=24mm, West=24mm\}$, at time $t_2$  $\{North=16mm, East=16mm,South=16mm, West=16mm\}$.
The observation map, which only relates data at  $t_0$ states that $\{North=5mm, East=5mm,South=5mm, West=5mm\}$.

We know from knowledge experts that ECMWF has a better accuracy than GFS: numerically $a(ECMWF,t_1)=0.85$, $a(ECMWF,t_2)=0.80$ , $a(GFS,t_1)=0.45$, $a(GFS,t_2)=0.40$

These assertions, using ``E'' for ECMWF,  ``G'' for GFS, ``O'' for observation and  ``R'' for ``rain'' can be represented as 

{\scriptsize 
\begin{tabular}{llll}
$\langle E,t_0 \rangle :R(North,t_0,4)$&$ \langle E,t_0 \rangle :R(East,t_0,4)$&$ \langle E,t_0 \rangle :R(South,t_0,4)$&$ \langle E,t_0 \rangle :R(West,t_0,4)$\\
$\langle E,t_0 \rangle :R(North,t_1,4)$&$ \langle E,t_0 \rangle :R(East,t_1,4)$&$ \langle E,t_0 \rangle :R(South,t_1,4)$&$ \langle E,t_0 \rangle :R(West,t_1,4)$\\
$\langle E,t_0 \rangle :R(North,t_2,6)$&$ \langle E,t_0 \rangle :R(East,t_2,6)$&$ \langle E,t_0 \rangle :R(South,t_2,6)$&$ \langle E,t_0 \rangle :R(West,t_2,6)$\\
$\langle G,t_0 \rangle :R(North,t_0,5)$&$ \langle G,t_0 \rangle :R(East,t_0,5)$&$ \langle G,t_0 \rangle :R(South,t_0,5)$&$ \langle G,t_0 \rangle :R(West,t_0,5)$\\
$\langle G,t_0 \rangle :R(North,t_1,24)$&$ \langle G,t_0 \rangle :R(East,t_1,24)$&$ \langle G,t_0 \rangle :R(South,t_1,24)$&$ \langle G,t_0 \rangle :R(West,t_1,24)$\\
$\langle G,t_0 \rangle :R(North,t_2,16)$&$ \langle E,t_0 \rangle :R(East,t_2,16)$&$ \langle G,t_0 \rangle :R(South,t_2,16)$&$ \langle G,t_0 \rangle :R(West,t_2,16)$\\
$\langle O,t_0 \rangle :R(North,t_0,5)$&$ \langle O,t_0 \rangle :R(East,t_0,5)$&$ \langle O,t_0 \rangle :R(South,t_0,5)$&$ \langle O,t_0 \rangle :R(West,t_0,5)$\\
\end{tabular}
}
~~\\
This is therefore our set of metarules, so after the \emph{Translator} has done its elaboration using algorithm described in \ref{algorithm} we can have:

{\scriptsize
\begin{tabular}{lllllllllll}
$r_{g_{11}}: $&$ \Rightarrow N_g{t_0}4$ &~~~~~~~~~~ &$r_{g_{21}}: $&$ \Rightarrow  N_g{t_1}4$&~~~~~~~~~~ &$r_{g_{31}}: $&$ \Rightarrow  N_g{t_2}6$ &~~~~~&$r_{o_{11}}: $&$ \rightarrow  N{t_0}5$\\
$r_{g_{12}}: $&$ \Rightarrow E_g{t_0}4$ & ~~~~~~~~~~ &$r_{g_{22}}: $&$ \Rightarrow  E_g{t_1}4$&~~~~~~~~~~ &$r_{g_{32}}: $&$ \Rightarrow  N_g{t_2}6$&~~~~~&$r_{o_{12}}: $&$ \rightarrow  E{t_0}5$ \\
$r_{g_{13}}: $&$ \Rightarrow S_g{t_0}4$ & ~~~~~~~~~~ &$r_{g_{23}}: $&$ \Rightarrow  S_g{t_1}4$&~~~~~~~~~~ &$r_{g_{33}}: $&$ \Rightarrow  N_g{t_2}6$&~~~~~&$r_{o_{13}}: $&$ \rightarrow  S{t_0}5$\\
$r_{g_{14}}: $&$ \Rightarrow W_g{t_0}4$ &~~~~~~~~~~  &$r_{g_{24}}: $&$ \Rightarrow  W_g{t_1}4$&~~~~~~~~~~ &$r_{g_{34}}: $&$ \Rightarrow  N_g{t_2}6$&~~~~~&$r_{o_{14}}: $&$ \rightarrow  W{t_0}5$ \\
 &&&&&&&&&&\\
$r_{g_{21}}: $&$ \Rightarrow N_g{t_1}4$ &~~~~~~~~~~ &$r_{e_{21}}: $&$ \Rightarrow  N_e{t_1}24$&~~~~~~~~~~ &$r_{e_{31}}: $&$ \Rightarrow  N_e{t_2}16$&&&\\
$r_{g_{22}}: $&$ \Rightarrow E_g{t_1}4$ & ~~~~~~~~~~ &$r_{e_{22}}: $&$ \Rightarrow  E_e{t_1}24$& ~~~~~~~~~~ &$r_{e_{32}}: $&$ \Rightarrow  E_e{t_2}16$&&&\\
$r_{g_{23}}: $&$ \Rightarrow S_g{t_1}4$ & ~~~~~~~~~~ &$r_{e_{23}}: $&$ \Rightarrow  S_e{t_1}24$& ~~~~~~~~~~ &$r_{e_{33}}: $&$ \Rightarrow  S_e{t_2}16$&&&\\
$r_{g_{24}}: $&$ \Rightarrow W_g{t_1}4$ &~~~~~~~~~~  &$r_{e_{24}}: $&$ \Rightarrow  W_e{t_1}24$&~~~~~~~~~~  &$r_{e_{34}}: $&$ \Rightarrow  W_e{t_2}16$&&&\\
% &&&&&&&&&&\\
\end{tabular}
 
\begin{tabular}{lllll}
$r_{fg_{11}}: $&$ N_g{t_1}4, N_e{t_1} 24 \Rightarrow N{t_1}7$ &~~~~~~~~~~ &$r_{fe_{11}}: $&$  N_g{t_1}4, N_e{t_1} 24 \Rightarrow  N{t_1}21$\\
$r_{fg_{12}}: $&$ E_g{t_1}4, E_e{t_1} 24 \Rightarrow E{t_1}7$ &~~~~~~~~~~ &$r_{fe_{12}}: $&$  E_g{t_1}4, E_e{t_1} 24 \Rightarrow  E{t_1}21$\\
$r_{fg_{13}}: $&$ S_g{t_1}4, S_e{t_1} 24 \Rightarrow S{t_1}7$ &~~~~~~~~~~ &$r_{fe_{13}}: $&$  S_g{t_1}4, S_e{t_1} 24 \Rightarrow  S{t_1}21$\\
$r_{fg_{14}}: $&$ W_g{t_1}4, W_e{t_1} 24 \Rightarrow W{t_1}7$ &~~~~~~~~~~ &$r_{fe_{14}}: $&$  W_g{t_1}4, W_e{t_1} 24 \Rightarrow  W{t_1}21$\\
&&&&\\
$r_{fg_{21}}: $&$ N_g{t_2}6, N_e{t_2} 16 \Rightarrow N{t_2}8$ &~~~~~~~~~~ &$r_{fe_{21}}: $&$  N_g{t_2}6, N_e{t_2} 16 \Rightarrow  N{t_2}14$\\
$r_{fg_{22}}: $&$ E_g{t_2}6, E_e{t_2} 16 \Rightarrow E{t_2}8$ &~~~~~~~~~~ &$r_{fe_{22}}: $&$  E_g{t_2}6, E_e{t_2} 16 \Rightarrow  E{t_2}14$\\
$r_{fg_{23}}: $&$ S_g{t_2}6, S_e{t_2} 16 \Rightarrow S{t_2}8$ &~~~~~~~~~~ &$r_{fe_{23}}: $&$  S_g{t_2}6, S_e{t_2} 16 \Rightarrow  S{t_2}14$\\
$r_{fg_{24}}: $&$ W_g{t_2}6, W_e{t_2} 16 \Rightarrow W{t_2}8$ &~~~~~~~~~~ &$r_{fe_{24}}: $&$  W_g{t_2}6, W_e{t_2} 16 \Rightarrow  W{t_2}14$\\
\end{tabular} 

\begin{tabular}{llllllllll}
%&&&&&&&&&\\
$v_{11}: $& $N{t_1}7  \Rightarrow \neg N{t_1}21 $ &~&$v_{21}: $&$N{t_1}21  \Rightarrow \neg N{t_1}7 $ &~~~$v_{31}: $& $N{t_2}8  \Rightarrow \neg N{t_2}14 $&~~~&$v_{41}: $&$N{t_2}14  \Rightarrow \neg N{t_2}8 $\\
$v_{12}: $& $E{t_1}7  \Rightarrow \neg E{t_1}21 $ &~&$v_{22}: $&$E{t_1}21  \Rightarrow \neg E{t_1}7 $  &~~~$v_{32}: $& $E{t_2}8  \Rightarrow \neg E{t_2}14 $&~~~&$v_{42}: $&$E{t_2}14  \Rightarrow \neg E{t_2}8 $\\
$v_{13}: $& $S{t_1}7  \Rightarrow \neg S{t_1}21 $ &~&$v_{23}: $&$S{t_1}21  \Rightarrow \neg S{t_1}7 $  &~~~$v_{33}: $& $S{t_2}8  \Rightarrow \neg S{t_2}14 $&~~~&$v_{43}: $&$S{t_2}14  \Rightarrow \neg S{t_2}8 $ \\
$v_{14}: $& $W{t_1}7  \Rightarrow \neg O{t_1}21 $ &~&$v_{24}: $&$O{t_1}21  \Rightarrow \neg O{t_1}7 $  &~~~$v_{34}: $& $W{t_2}8  \Rightarrow \neg O{t_2}14 $&~~&$v_{44}: $&$O{t_2}14  \Rightarrow \neg O{t_2}8 $ \\
%&&&&&&&&&\\
\end{tabular} 

\begin{tabular}{lllllllllll}
$p_{11}: $& $ v_{21}   ~~~ \rangle ~~~  r_{fg_{11}} $&~~~~~~~~~~ &$p_{21}: $&$ r_{fe_{11}} ~~~ \rangle ~~~  v_{11} $ &~~~ &$p_{31}: $& $ v_{41}   ~~~ \rangle ~~~  r_{fg_{21}} $&~~~~~~~~~~ &$p_{41}: $&$ r_{fe_{21}} ~~~ \rangle ~~~  v_{31} $\\
$p_{12}: $& $ v_{22}   ~~~ \rangle ~~~  r_{fg_{12}} $&~~~~~~~~~~ &$p_{22}: $&$ r_{fe_{12}} ~~~ \rangle ~~~ v_{12} $ &~~~ &$p_{32}: $& $ v_{42}   ~~~ \rangle ~~~  r_{fg_{22}} $&~~~~~~~~~~ &$p_{42}: $&$ r_{fe_{22}} ~~~ \rangle ~~~ v_{32} $\\
$p_{13}: $& $ v_{23}   ~~~ \rangle ~~~  r_{fg_{13}} $&~~~~~~~~~~ &$p_{23}: $&$ r_{fe_{13}} ~~~ \rangle ~~~ v_{13} $ &~~~& $p_{33}: $& $ v_{43}   ~~~ \rangle ~~~  r_{fg_{23}} $&~~~~~~~~~~ &$p_{44}: $&$ r_{fe_{23}} ~~~ \rangle ~~~ v_{33} $\\
$p_{14}: $& $ v_{24}   ~~~ \rangle ~~~  r_{fg_{14}} $&~~~~~~~~~~ &$p_{24}: $&$ r_{fe_{14}} ~~~ \rangle ~~~ v_{14} $ &~~~& $p_{34}: $& $ v_{44}   ~~~ \rangle ~~~  r_{fg_{24}} $&~~~~~~~~~~ &$p_{44}: $&$ r_{fe_{24}} ~~~ \rangle ~~~ v_{34} $\\
\end{tabular} 
}
\hspace{1cm}\\

Given that theory, the \emph{Reasoner} concludes $+\partial N{t_1}21$, $+\partial E{t_1}21$, $+\partial S{t_1}21$, $+\partial W{t_1}21$, $+\partial ~\nega N{t_1}7$, $+\partial ~\nega E{t_1}7$, $+\partial ~\nega S{t_1}7$, $+\partial ~\nega W{t_1}7$,        $+\partial N{t_2}14$, $+\partial E{t_2}14$, $+\partial S{t_2}14$, $+\partial W{t_2}14$, $+\partial ~\nega N{t_2}8$, $+\partial ~\nega E{t_2}8$, $+\partial ~\nega S{t_2}8$, $+\partial ~\nega W{t_2}8$.

Given that, translating numerical value into words, we have at $t_1$ that $\{$ North=Heavy Rain, East=Rain,South=Heavy Rain, West=Heavy Rain $\}$ while at time $t_2$  $\{$ North=Strong Rain, East=Strong Rain,South=Possible Showers, West=Possible Showers $\}$ which translates in the following figures.

\begin{tabular}{cc}

\includegraphics[width=5cm]{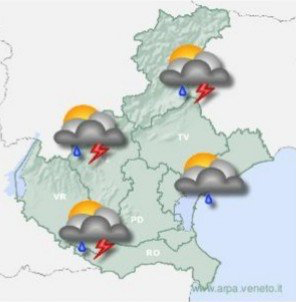}&\includegraphics[width=5cm]{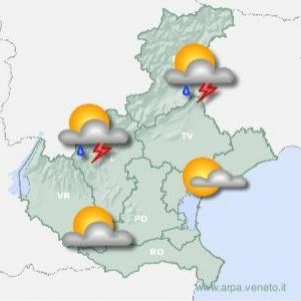}\\
Weather forecats at $t_1$ &Weather forecats at $t_2$\\ 
\end{tabular} 

%% file: SPINDLE_CONCLUSIONSTWO.tex
\subsection{Spindle conclusions for rules of the reference implementation}
\begin{verbatim}
*******************************************
* SPINdle (version 2.2.4)                       
* Copyright (C) 2009-2013 NICTA Ltd.       
......
* java -jar spindle-<version>.jar --app.license  
****************************************************
=========================
== application start!! ==
=========================
Initialize application context - start
.....
===================
......
+d E_t0_5(X)
    +d E_t1_14(X)
    +d E_t1_21(X)
    +d -E_t1_7(X)
    +d E_t1_8(X)
    +d Ee_t0_5(X)
    +d Ee_t1_24(X)
    +d Ee_t2_16(X)
    +d Eg_t0_4(X)
    +d Eg_t1_4(X)
    +d Eg_t2_6(X)
    +d N_t0_5(X)
    +d N_t1_14(X)
    +d N_t1_21(X)
    +d -N_t1_7(X)
    +d N_t1_8(X)
    +d Ne_t0_5(X)
    +d Ne_t1_24(X)
    +d Ne_t2_16(X)
    +d Ng_t0_4(X)
    +d Ng_t1_4(X)
    +d Ng_t2_6(X)
    +d S_t0_5(X)
    +d S_t1_14(X)
    +d S_t1_21(X)
    +d -S_t1_7(X)
    +d S_t1_8(X)
    +d Se_t0_5(X)
    +d Se_t1_24(X)
    +d Se_t2_16(X)
    +d Sg_t0_4(X)
    +d Sg_t1_4(X)
    +d Sg_t2_6(X)
    +d W_t0_5(X)
    +d W_t1_14(X)
    +d W_t1_21(X)
    +d -W_t1_7(X)
    +d W_t1_8(X)
    +d We_t0_5(X)
    +d We_t1_24(X)
    +d We_t2_16(X)
    +d Wg_t0_4(X)
    +d Wg_t1_4(X)
    +d Wg_t2_6(X)
    -d -E_t1_14(X)
    -d -E_t1_21(X)
    -d E_t1_7(X)
    -d -E_t1_8(X)
    -d E_t2_14(X)
    -d E_t2_8(X)
    -d -N_t1_14(X)
    -d -N_t1_21(X)
    -d N_t1_7(X)
    -d -N_t1_8(X)
    -d N_t2_14(X)
    -d N_t2_8(X)
    -d -S_t1_14(X)
    -d -S_t1_21(X)
    -d S_t1_7(X)
    -d -S_t1_8(X)
    -d S_t2_14(X)
    -d S_t2_8(X)
    -d -W_t1_14(X)
    -d -W_t1_21(X)
    -d W_t1_7(X)
    -d -W_t1_8(X)
    -d W_t2_14(X)
    -d W_t2_8(X)

Calling the shutdown routine...
Terminate application context - start
Terminate application context - end
=======================================
=== Application shutdown completed! ===
=======================================
\end{verbatim}

%% file: main.bbl
\begin{thebibliography}{10}

\bibitem{Agresta2014417}
A.~Agresta, G.~Fattoruso, M.~Pollino, F.~Pasanisi, C.~Tebano, S.~De~Vito, and
  G.~Di~Francia.
\newblock An ontology framework for flooding forecasting.
\newblock {\em Lecture Notes in Computer Science (including subseries Lecture
  Notes in Artificial Intelligence and Lecture Notes in Bioinformatics)}, 8582
  LNCS(PART 4):417--428, 2014.

\bibitem{AZ13}
F.~Aschieri and M.~Zorzi.
\newblock Non-determinism, non-termination and the strong normalization of
  system t.
\newblock {\em Lecture Notes in Computer Science (including subseries Lecture
  Notes in Artificial Intelligence and Lecture Notes in Bioinformatics)}, 7941
  LNCS:31--47, 2013.

\bibitem{AZ16}
F.~Aschieri and M.~Zorzi.
\newblock {On natural deduction in classical first-order logic: Curry-Howard
  correspondence, strong normalization and Herbrand's theorem}.
\newblock {\em Theoretical Computer Science}, 625:125--146, 2016.

\bibitem{CarrIII2001355}
L.~Carr~III, R.~Elsberry, and J.~Peak.
\newblock Beta test of the systematic approach expert system prototype as a
  tropical cyclone track forecasting aid.
\newblock {\em Weather and Forecasting}, 16(3):355--368, 2001.

\bibitem{Conway198923}
B.~Conway.
\newblock Expert systems and weather forecasting.
\newblock {\em Meteorological Magazine}, 118(1399):23--30, 1989.

\bibitem{DBLP:conf/kes/CristaniDOTZ18}
M.~Cristani, F.~Domenichini, F.~Olivieri, C.~Tomazzoli, and M.~Zorzi.
\newblock It could rain: weather forecasting as a reasoning process.
\newblock In R.~J. Howlett, L.~C. Jain, Z.~Popovic, D.~B. Popovic, S.~N.
  Vukosavic, C.~Toro, and Y.~Hicks, editors, {\em Knowledge-Based and
  Intelligent Information {\&} Engineering Systems: Proceedings of the 22nd
  International Conference KES-2018, Belgrade, Serbia, 3-5 September 2018.},
  volume 126 of {\em Procedia Computer Science}, pages 850--859. Elsevier,
  2018.

\bibitem{cotz18}
M.~Cristani, F.~Olivieri, C.~Tomazzoli, and M.~Zorzi.
\newblock Towards a logical framework for diagnostic reasoning.
\newblock In {\em Agents and Multi-Agent Systems: Technologies and Applications
  2018. Proc. of KES-AMSTA-18, https://doi.org/10.1007/978-3-319-92031-3\_14},
  pages 1--12. Springer, 2018.

\bibitem{claudio16}
M.~Cristani, C.~Tomazzoli, E.~Karafili, and F.~Olivieri.
\newblock Defeasible reasoning about electric consumptions.
\newblock In {\em 30th {IEEE} International Conference on Advanced Information
  Networking and Applications, {AINA} 2016, Crans-Montana, Switzerland, 23-25
  March, 2016}, pages 885--892, 2016.

\bibitem{Desmarais1990150}
M.~C. Desmarais, F.~de~Verteuil, P.~Zwack, and D.~Jacob.
\newblock Stratus: A prototype expert advisory system for terminal weather
  forecasting.
\newblock pages 150--154, 1990.

\bibitem{Domanska20127673}
D.~Doma\'nska and M.~Wojtylak.
\newblock Application of fuzzy time series models for forecasting pollution
  concentrations.
\newblock {\em Expert Systems with Applications}, 39(9):7673--7679, 2012.

\bibitem{Domanska201419}
D.~Doma\'nska and M.~Wojtylak.
\newblock Explorative forecasting of air pollution.
\newblock {\em Atmospheric Environment}, 92:19--30, 2014.

\bibitem{Fohn199819}
P.~F\"ohn.
\newblock An overview of avalanche forecasting models and methods.
\newblock {\em Publikasjon - Norges Geotekniske Institutt}, (203):19--27, 1998.

\bibitem{Goldberg1993156}
E.~Goldberg.
\newblock Fog. synthesizing forecast text directly from weather maps.
\newblock pages 156--162, 1993.

\bibitem{Governatori2016230}
G.~Governatori, F.~Olivieri, E.~Calardo, A.~Rotolo, and M.~Cristani.
\newblock Sequence semantics for normative agents.
\newblock {\em Lecture Notes in Computer Science (including subseries Lecture
  Notes in Artificial Intelligence and Lecture Notes in Bioinformatics)}, 9862
  LNCS:230--246, 2016.

\bibitem{Governatori2014168}
G.~Governatori, F.~Olivieri, S.~Scannapieco, and M.~Cristani.
\newblock The hardness of revising defeasible preferences.
\newblock {\em Lecture Notes in Computer Science (including subseries Lecture
  Notes in Artificial Intelligence and Lecture Notes in Bioinformatics)}, 8620
  LNCS:168--177, 2014.

\bibitem{Governatori2016296}
G.~Governatori, F.~Olivieri, S.~Scannapieco, A.~Rotolo, and M.~Cristani.
\newblock The rationale behind the concept of goal.
\newblock {\em Theory and Practice of Logic Programming}, 16(3):296--324, 2016.

\bibitem{Hansen199759}
B.~Hansen.
\newblock Sigmar: a fuzzy expert system for critiquing marine forecasts.
\newblock {\em AI Applications}, 11(1):59--68, 1997.

\bibitem{Kulkarni2018555}
A.~Kulkarni and D.~Mukhopadhyay.
\newblock Internet of things based weather forecast monitoring system.
\newblock {\em Indonesian Journal of Electrical Engineering and Computer
  Science}, 9(3):555--557, 2018.

\bibitem{Kumar1994373}
V.~Kumar, C.~Chung, and C.~Lindley.
\newblock Toward building an expert system for weather forecasting operations.
\newblock {\em Expert Systems With Applications}, 7(2):373--381, 1994.

\bibitem{Mahabir20033749}
C.~Mahabir, F.~Hicks, and A.~Fayek.
\newblock Application of fuzzy logic to forecast seasonal runoff.
\newblock {\em Hydrological Processes}, 17(18):3749--3762, 2003.

\bibitem{Marra1996475}
R.~M. Marra, D.~H. Jonassen, and P.~Knight.
\newblock Use of expert system generation to promote knowledge synthesis in a
  meteorology forecasting course.
\newblock pages 475--479, 1996.

\bibitem{MVZ-jmvl}
A.~Masini, L.~Vigan{\`o}, and M.~Zorzi.
\newblock Modal deduction systems for quantum state transformations.
\newblock {\em J. Mult.-Valued Logic Soft Comput.}, 17(5-6):475--519, 2011.

\bibitem{McCarthy1990228}
J.~McCarthy and R.~J. Serafin.
\newblock Advanced aviation weather system based on new weather sensing
  technologies.
\newblock pages 228--239, 1990.

\bibitem{Moninger1990457}
W.~Moninger.
\newblock Shootout-89, an evaluation of knowledge-based weather forecasting
  systems.
\newblock {\em Machine Intelligence and Pattern Recognition}, 10(C):457--458,
  1990.

\bibitem{Olivieri2013213}
F.~Olivieri, G.~Governatori, S.~Scannapieco, and M.~Cristani.
\newblock Compliant business process design by declarative specifications.
\newblock {\em Lecture Notes in Computer Science}, 8291 LNAI:213--228, 2013.

\bibitem{Prawitz65}
D.~Prawitz.
\newblock Ideas and results in proof theory.
\newblock In J.~Fenstad, editor, {\em Proceedings of the Second Scandinavian
  Logic Symposium}, volume~63 of {\em Studies in Logic and the Foundations of
  Mathematics}, pages 235 -- 307. Elsevier, 1971.

\bibitem{Ramos-Soto201544}
A.~Ramos-Soto, A.~Bugarín, S.~Barro, and J.~Taboada.
\newblock Linguistic descriptions for automatic generation of textual
  short-term weather forecasts on real prediction data.
\newblock {\em IEEE Transactions on Fuzzy Systems}, 23(1):44--57, 2015.

\bibitem{Tomazzoli2017345}
C.~Tomazzoli, M.~Cristani, E.~Karafili, and F.~Olivieri.
\newblock Non-monotonic reasoning rules for energy efficiency.
\newblock {\em Journal of Ambient Intelligence and Smart Environments},
  9(3):345--360, 2017.

\bibitem{ViganoVZ14}
L.~Vigan{\`{o}}, M.~Volpe, and M.~Zorzi.
\newblock Quantum state transformations and branching distributed temporal
  logic - (invited paper).
\newblock In {\em Logic, Language, Information, and Computation - 21st
  International Workshop, WoLLIC 2014, Valpara{\'{\i}}so, Chile, September 1-4,
  2014. Proceedings}, pages 1--19, 2014.

\bibitem{ViganoVZ17}
L.~Vigan{\`{o}}, M.~Volpe, and M.~Zorzi.
\newblock A branching distributed temporal logic for reasoning about
  entanglement-free quantum state transformations.
\newblock {\em Inf. Comput.}, 255:311--333, 2017.

\end{thebibliography}
